%% file: main.tex
\documentclass[10pt,compsoc]{IEEEtran}
\usepackage[labelfont=bf,textfont={bf}]{caption}
\usepackage{textcomp}
\usepackage{dblfloatfix}  

\usepackage{lastpage}
\usepackage{lineno}
\usepackage{balance}
\usepackage{hyperref}
\usepackage{picture}
\usepackage{relsize}
\usepackage{multicol}
\usepackage{soul}
\usepackage{enumitem}
\setitemize{noitemsep,topsep=0pt,parsep=0pt,partopsep=0pt,leftmargin=*}

\input{macros.tex}
\newcommand{\IT}{\sffamily{FairMask}}
\newcommand{\bi}{\begin{itemize}}
\newcommand{\ei}{\end{itemize}}
\renewcommand{\footnotesize}{\scriptsize}
\usepackage{booktabs}
\usepackage[ruled,vlined]{algorithm2e}
\newtcolorbox{blockquote}{colback=blue!5,boxrule=0.4pt,colframe=black,fonttitle=\bfseries}

\newcommand{\BLACK}{\color{black}}
\definecolor{black}{rgb}{0,0,0}
\definecolor{Blue}{RGB}{0,29,193}

\begin{document}

\title{{\IT}: Better Fairness via Model-based Rebalancing of Protected Attributes }

\author{Kewen~Peng,  Joymallya Chakraborty
        Tim~Menzies,~\IEEEmembership{Fellow,~IEEE}
\IEEEcompsocitemizethanks{\IEEEcompsocthanksitem K. Peng, J. Chakraborty and  T. Menzies are with the Department
of Computer Science, North Carolina State University, Raleigh, USA.
 \protect
E-mail:kpeng@ncsu.edu, jchakra@ncsu.edu, timm@ieee.org}}

\markboth{IEEE Transactions on Software Engineering}%
{Peng \MakeLowercase{\textit{et al.}}: FairMask for IEEE Journals}

\IEEEtitleabstractindextext{
\begin{abstract}
\textbf{Context}: Machine learning software can generate models that inappropriately  discriminate against specific protected social groups (e.g., groups based on gender, ethnicity, etc.). Motivated by those results, software engineering researchers have proposed many methods for mitigating  those discriminatory effects.
While those methods are effective in mitigating bias, few of them can provide explanations on what is the root cause of bias.

\noindent
\textbf{Objective}: We aim to better detect and mitigate algorithmic discrimination in machine learning software problems. 

\noindent
\textbf{Method}: Here we propose {\IT}, a {\em model-based} extrapolation method that is capable of both mitigating bias and explaining the cause.
In our {\IT} approach, protected attributes are represented by models learned from the other independent variables (and these  models  offer extrapolations over the space between existing examples). We then use the extrapolation models to relabel protected attributes later seen in testing data or deployment time. Our approach aims to offset the biased predictions of the classification model via rebalancing the distribution of protected attributes.

\noindent
\textbf{Results}:The experiments of this paper show that, without compromising (original) model performance, {\IT} can achieve significantly better group and individual fairness (as measured in different metrics) than benchmark methods. Moreover, compared to another instance-based rebalancing method, our model-based approach shows faster runtime and thus better scalability.

\noindent
\textbf{Conclusion}: 
Algorithmic decision bias can be removed via extrapolation that smooths away   outlier points. As evidence for this, 
our proposed {\IT} is not
only performance-wise better (measured by
fairness and performance metrics) than two state-of-the-art fairness algorithms.

\noindent
\textbf{Reproduction Package}:In order to better support open science, all scripts and data used in this study are available on-line at
 \url{https://github.com/anonymous12138/biasmitigation}.

\end{abstract}

\begin{IEEEkeywords}
Software Fairness, Explanation, Bias Mitigation
\end{IEEEkeywords}}

\maketitle
\IEEEpeerreviewmaketitle
\IEEEdisplaynontitleabstractindextext

\section{Introduction}\label{intro}

Increasingly, machine learning (ML) algorithms are applied in software engineering (SE) to assist decision-making, and some of the decisions take private information (e.g., race, gender, age) of human individuals into consideration. For example, ML models are used in software to assist determine which loan applications should get approved; which citizens should get bail; which patients can be released from the hospital.
From an ethical perspective, using private information also puts such software under the exposure to unintentionally algorithmic discrimination, where the benefits of certain social groups are compromised.
Many prior cases have shown the existence of such flaws: Google's sentimental analysis model was found to assign negative scores to homosexual or Jewish attributes in a sentence; In machine translation, translators wrongly relabel doctors as male and nurses as female; In credit card applications, applicants with similar conditions are receiving significantly different credit lines based on their genders~\cite{DEFAULT}.    



Many researchers are endeavoring to resolve the discrimination issue in ML software. Recent success with the Fair-SMOTE~\cite{Chakraborty2021BiasIM} of Chakraborty et al. shows that it is possible to carefully rebalance the training data in order to mitigate bias in the data.  
Chakraborty et al. conjectured that models are unfair when the training data does not equally represent all social groups.  Fair-SMOTE uses a rebalancing method that adjusts the training data such that all values of "protected attributes" are equally represented in the training data
(and by ``protected attributes'' we mean information
about age, gender, racial origins, veteran status, etc. 
that is used to identify a person as belonging
to corresponding groups, some of which
have suffered from social injustice in history).


While a successful system in its test domain,
Fair-SMOTE has problems with 
{\em procedural justice}.
By definition~\cite{tyler2002procedural,lee2019procedural}, procedural justice requires not only  fair results but also transparency of the decision-making process such that ones can verify whether the procedure guarantees fairness. One way to demonstrate procedural justice to
a group of users is to ensure that 
an AI system never asks about protected attributes.
Note that this is {\em not} the case with Fair-SMOTE
since when its models are deployed, all the new examples
must have the same format as the data seen during training.
Therefore,  the protected attributes data must be collected from users.
Consequently, users can grow concerned that the model
will not mitigate against bias (since it has access to
the protected information).

Accordingly, this paper explores an
alternative that, once developed, no longer needs
to access protected attributes during real-time deployment.
In our concept of operation,
our method only collects and uses
those protected attributes to initially build its model
(which are assessed using widely accepted fairness metrics, see Table~\ref{tab:metrics}).
After that, during deployment, our method does
not demand access to protected attributes in subsequent test data.  

Removing protected attributes must be done carefully.
Prior research has shown that it is insufficient to just remove projected attributes from data.
If a model merely ignores the protected attributes,
then that can
either (a) harm the performance of the prediction model due to information loss~\cite{NIPS2017_6988,mehrabi2021survey}, or (b) have a trivial influence on improving group fairness due to proxy discrimination~\cite{kamiran2012data,kamiran2009classifying,kilbertus2017avoiding}.
Bias can persist due to the correlations
between variables. Such correlations indicate
that an unwanted bias can persist
(even though the protected attribute is removed).
For an example of proxy discrimination,
consider how residential zip codes can be used to make biased decisions such as granting loans since zip codes might correlate to race given historical causes.~\cite{zhang2016causal}.

To address these issues, our {\IT} works as follows:
\bi
\item The algorithm will avoid inferring protected attributes after a model is initially built.
\item For new incoming data instances, it then artificially masks those protected attributes via an {\em extrapolation model} learned from other non-protected attributes.
\ei
As shown by the results of this paper,
{\IT} shows on-par or superior performance compared to the prior state-of-the art (Reweighing and Fair-SMOTE):
\bi
\item {\IT} provided better bias reduction with as good or better predictive performance;
\item {\IT} runs much faster  (up to 600\%) than Fair-SMOTE;
\item {\IT} scales better to larger data sets;
\item {\IT} handles multiple protected attributes very well (and the Fair-SMOTE paper notes that managing multiple attributes is an Achilles's heel of that algorithm).
\ei
Importantly, {\IT} ensures procedural justice. {\IT}
needs to access protected attributes during the initial commission stage, but not during deployment. As a result,
when this system is placed into production, it needs no access to users' private information when making a prediction.
 
The rest of this paper is structured as follows. \S\ref{rw} provides a road-map of background knowledge and related work concerning fairness in ML software. \S\ref{explain} describes the motivation and methodology of our approach in this paper. \S\ref{experiment} illustrates the experiment setup used to evaluate our approach along with other benchmarks. \S\ref{result} shows experiment results. \S\ref{threat} lists external and internal threats to validity in this paper. In \S\ref{discussion}, we elaborate the reasons why FairMask should be promoted.  Finally, \S\ref{conclusion} presents our conclusions.
 
\begin{table*}[t!]
\centering
\scriptsize
\caption{Description of datasets used in this paper.}
\begin{tabular}{ccclcc}
\toprule
Dataset &
  \#Features &
  \#Rows &
  \multicolumn{1}{c}{Domain} &
  Protected Attribute &
  Favorable Label \\
  \midrule
Adult Census~\cite{ADULT} & 14  & 48,842 & U.S. census information from 1994 to predict personal income & Sex, Race &
  Income $>$ \$50,000 \\
Compas~\cite{COMPAS}         & 28    & 7,214  & Criminal history of defendants to predict re-offending         & Sex, Race & Re-offend $=$ false   \\
German Credit~\cite{GERMAN}  & 20    & 1,000  & Personal information to predict good or bad credit                              & Sex       & Credit $=$ good   \\
Bank Marketing~\cite{BANK} & 16    & 45,211 & Marketing data of a Portuguese bank to predict term deposit                     & Age       & Subscription $=$ yes \\
Heart Health~\cite{HEART}   & 14    & 297    & Patient information from Cleveland DB to predict heart disease                  & Age       & Diagnose $=$ yes   \\
Default Credit~\cite{DEFAULT} & 23   & 30,000 & Customer information in Taiwan to predict default payment         & Sex       & Payment $=$ yes   \\
MEPS15~\cite{MEPS}        & 1831  & 4,870 & Surveys of household members and their medical providers & Race      & Utilization $>=$ 10  \\
\bottomrule
\end{tabular}
\label{tab:dataset}
\end{table*}

\begin{table*}[t!]
\centering
\scriptsize
\caption{Definitions and descriptions of fairness metrics used in this paper.}

\begin{tabular}{|l|l|l|}
\hline
\multicolumn{1}{c}{Metric} &
  \multicolumn{1}{c}{Definition} &
 \multicolumn{1}{c}{Description} \\ \hline
Average Odds Difference (AOD) &
  \begin{tabular}[c]{@{}l@{}}TPR $=$ TP$/$(TP $+$ FN), FPR $=$ FP/(FP $+$ TN)\\ AOD= (($FPR_{U} - FPR_{P}$) + ($TPR_{U} - TPR_{P}$))$/$2\end{tabular} &
  \begin{tabular}[c]{@{}l@{}}Average of difference in False Positive Rates(FPR) and True\\ Positive Rates(TPR) for unprivileged and privileged groups\end{tabular} \\ \hline
Equal Opportunity Difference (EOD) &
  EOD $=$ $TPR_{U} - TPR_{P}$ &
  \begin{tabular}[c]{@{}l@{}}Difference of True Positive Rates(TPR) for unprivileged and\\ privileged groups\end{tabular} \\ \hline
Statistical Parity Difference (SPD) &
  SPD $=$ P (Y $=$ 1$|$PA $=$ 0) $-$ P (Y $=$ 1$|$PA $=$ 1) &
  \begin{tabular}[c]{@{}l@{}}Difference between probability of unprivileged group \\ (protected attribute PA $=$ 0) gets favorable prediction (Y $=$ 1)\\ \& probability of privileged group (protected attribute PA $=$ 1)\\ gets favorable prediction (Y $=$ 1)\end{tabular} \\ \hline
Disparate Impact (DI) &
  DI $=$ P {(}Y $=$ 1|PA $=$ 0{]}$/$P {[}Y $=$ 1|PA $=$ 1{)} &
  Similar to SPD but measuring ratio rather than the probability \\ \hline
Flip Rate (FR) &
  FR $=$ $\Sigma$(L$|$L{[}PA$=$0{]} $\not =$ L{[}PA$=$1{]})$/ total$ &
  \begin{tabular}[c]{@{}l@{}}The ratio of instances whose predicted label ($L$) will change\\ when flipping their protected attributes (e.g., PA$=$1 to PA$=$0) \end{tabular} \\ \hline
\end{tabular}
\label{tab:metrics}
\end{table*}

\section{Background and Related Work}\label{rw}
In this section, we introduce fundamental theories about software fairness, metrics to measure it, and related works that attempt to mitigate bias. 

\subsection{Why Software Engineers Care About Fairness}

The rapid development of ML has greatly benefited SE practitioners, and examples of ML-assisted software can be found everywhere: defect prediction models used to locate the most error-prone code files in the upcoming releases; effort estimations tools used to better manage human and capital resources; multi-objective optimizers used to generate configuration solutions for system of enormous configurable options. Meanwhile, ethical concerns have also drawn increasing attention in the ML and SE communities. 

While in many scenarios, the only utility needed to be optimized is the performance of the models (in tasks about prediction, classification, ranking, etc.), other cases where private information of human beings is collected need to be handled more carefully. ML software systems have been deployed in many areas to assist make decisions that affect human individuals:  Courts and corrections departments in the US use software to determine sentence length for defendants~\cite{feller2016computer}; algorithms are used to predict the default payments from credit card users~\cite{yeh2009comparisons}. During such procedures, private information such as age, ethnicity, and gender are collected.
Moreover, it has been revealed in prior studies that models learned from such data may contain algorithmic bias toward certain social groups.

In response to the above raising issues, IEEE has provoked ethical designs of AI-assisted systems~\cite{shahriari2017ieee} and the European Union also announced the ethics guidelines for building trustworthy AI~\cite{doi/10.2759/177365}. Fairness has been emphasized in both documents. Big industrial companies such as Facebook~\cite{Fairness_Flow}, Microsoft~\cite{FATE}, and Google~\cite{simonite2020google} also have begun to invest effort in ensuring fairness of their products. In academia, IEEE and ACM have set specific tracks~\cite{FAT,EXPLAIN} for papers studying fairness problems.

\subsection{Fairness in ML Software}
In this work, we study binary classification problems. We define some terms specific to the fairness of binary classification. 
\begin{itemize}
    \item A {\em favorable label} in a binary classification task is the label that grants the instance (usually human individuals) with privilege such as a job offer or being accepted for a loan.
    \item A {\em protected/sensitive attribute} reveals the social groups to which data instances belong, such as gender, race, and age. A binary protected attribute will divide the whole population into {\em privileged} and {\em unprivileged} groups in terms of the difference in receiving the favorable label. 
\end{itemize}
The notion of bias rises if the outcome of the classification model gets significantly affected by protected/protected attributes. Table~\ref{tab:dataset} shows seven fairness datasets used in this work. These datasets are very popular in the fairness domain and have been widely used by many prior researchers~\cite{Chakraborty_2020,Chakraborty2021BiasIM,chakraborty2019software,9286091,Biswas_2020}. All of these datasets contain at least one protected attribute. Depending on that, the population is divided into two groups getting different benefits. For example, in the Adult~\cite{ADULT} dataset, there are two protected attributes. Based on ``sex'', ``male'' is privileged; Based on ``race'', ``white'' is privileged.

The concept of fairness is complicated and very domain-specific. Narayanan~\cite{Arvind} has defined 21 different versions of fairness. Based on prior literature~\cite{Biswas_2020,Chakraborty_2020,Chakraborty2021BiasIM}, among these 21 versions, two specific versions of fairness are widely explored and given most importance. We have decided to explore the same two versions and chose different metrics to evaluate them. 

\begin{itemize}
    \item {\em Group fairness} requires the approximate equalization
    of certain statistical property across groups divided by the protected attribute. In this paper, we use 4 group fairness metrics that were widely used in prior research~\cite{kamiran2012data,feldman2015certifying,Chakraborty_2020,Chakraborty2021BiasIM,Biswas_2020}.
    \item {\em Individual fairness} requires that similar individuals should receive similar prediction outcomes by the ML model. The usual metric for measuring individual fairness is ``consistency''. But ``consistency'' is a collective metric based on nearest neighbors. That means it can be calculated for a set of data points, not for a single point. Chakraborty et al.~\cite{Chakraborty_2020} introduced a new metric for measuring individual fairness where they measured the ratio of the population whose prediction outcomes are flipped (e.g., accepted to rejected) when reversing their protected attributes. We decided to use the same metric called {\it Flip Rate} (FR).
\end{itemize}
Table~\ref{tab:metrics} contains mathematical definitions of all 5 fairness metrics. All the group fairness metrics are calculated based on the confusion matrix of binary classification, which is consisted of four parts: 
true positive (TP), true negative (TN), false positive (FP), and false negative (FN).

\subsection{Bias Mitigation}\label{bias-mitigation}
Many researchers endeavor to ensure
fairness within their AI decision-making software.
From the literature, one can categorize bias mitigation methods into three major groups, depending on when the mitigation procedure is performed.\\
 \\
    \noindent
    {\bf Pre-processing}: Pre-processing algorithms attempt to mitigate the bias of the classification model by pre-processing the training data that a model learns from. Reweighing was proposed by Kamiran et al.~\cite{kamiran2012data} to learn a probabilistic threshold that can generate weights to different instances in training samples according to the (protected and class attributes) combination that each of them belongs to. Fair-SMOTE~\cite{Chakraborty2021BiasIM} proposed by Chakraborty et al. re-samples and generates synthetic instances among the training data so that the training data can reach equal distributions not only between different target labels but also among different protected attributes. SRCVAE, a more recent algorithm by Grari et al., uses auto-encoding to generate a sensitive information proxy such that the protected attribute will not be required when training a model~\cite{grari2021fairness}. \\
    \\
    \noindent
    {\bf In-processing}: In-processing methods generally take the optimization approach to mitigate bias. The dataset is typically divided into three parts: training, validation, and testing set. The learner is fitted on the training set and then optimized on the validation set using both performance and fairness metrics as the objectives. Kamishima et al.~\cite{10.1007/978-3-642-33486-3_3} developed Prejudice Remover which adds a discrimination-aware regularization to the learning objective of the model. Other recent works also attempt using ensemble learning or multi-task learning to tackle the fairness-performance optimization problem~\cite{oneto2019taking,chen2022maat}. \BLACK\\
    \\
    \noindent
    {\bf Post-processing}: This category of approaches generally believe that bias can be removed by identifying and then reversing the biased outcomes from the classification model, which means that such methods typically only modify outcomes of the model rather than the model itself. The  ``reject option classification'' (ROC) approach was proposed by Kamiran et al.~\cite{kamiran2012decision,Kamiran:2018:ERO:3165328.3165686} to identify the model's decision boundary with the highest uncertainty. Within that region, the method will separately adjust the classification thresholds for (a) favorable labels on unprivileged groups and (b) unfavorable labels on privileged groups.\\
\\

This paper works under the same assumptions claimed by 
Chakraborty et al~\cite{Chakraborty2021BiasIM}. That is, one of the major root causes of bias is imbalanced training data. However, different from Fair-SMOTE, which is a pre-processing method, we approach the problem via a hybrid of pre-processing and post-processing methods. Our advantages over a pre-processing method are: (a) Because FairMask does not pre-process the training data, it does not require re-training an already-built model; (b) Hence FairMask can be added to a model during deployment with little effort and cost. 
\section{Methodology}\label{explain}  

This section illustrates the designs of {\IT}. Before introducing our approach, we present two essential conjectures made in this paper and the follow-up discussion. Our design of {\IT} is based on these conjectures.
\begin{figure*}[b!]
\centering
\includegraphics[width=.99\textwidth]{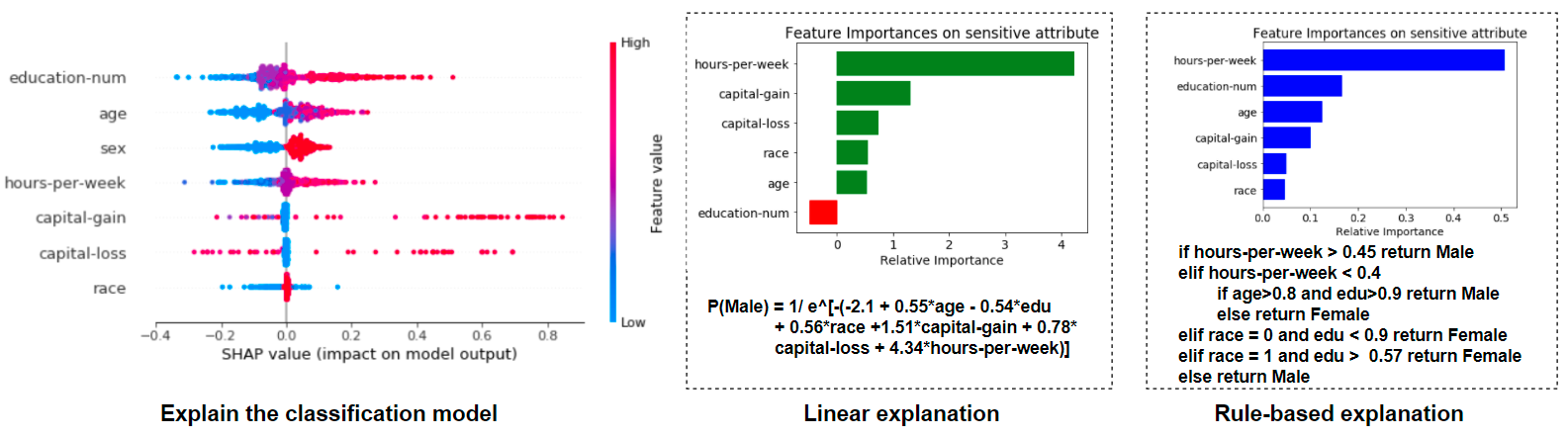}
\caption{
Example based on the Adult Income dataset. The left side is the explanation on the dependent attribute, in form of SHAP explanations~\cite{lundberg2017unified} of the classification model. The middle and right blocks show two approaches (logistic regression and decision tree, respectively) applied in FairMask to explain the influence of other independent attributes on the protected attribute.  } 
\label{fig:example1}
\end{figure*}
\subsection{Explaining Bias}\label{xplanb}
Much prior work has searched for a root cause of bias.
Creager et al.~\cite{creager2019flexibly} proposed to use disentangled representation learning to identify potential bias-introducing latent in training data that contains mutual information of both targets and protected attributes. They then add regularization on the mutual information while also optimizing for the predictive power.
Similarly, Park et al.~\cite{park2020readme} proposed to disentangle information of the target attribute and protected attribute such that target-related information is preserved while protected-related information is removed. It is noteworthy that while both works were empirically tested effective in bias mitigation, the neural-network-based disentanglement approach is barely interpretable, which means the internal disentanglement process cannot be presented to users in a comprehensible manner. We view this as a transparency issue and propose an alternative approach.
\begin{figure*}[b!]
\centering
\includegraphics[width=.90\textwidth]{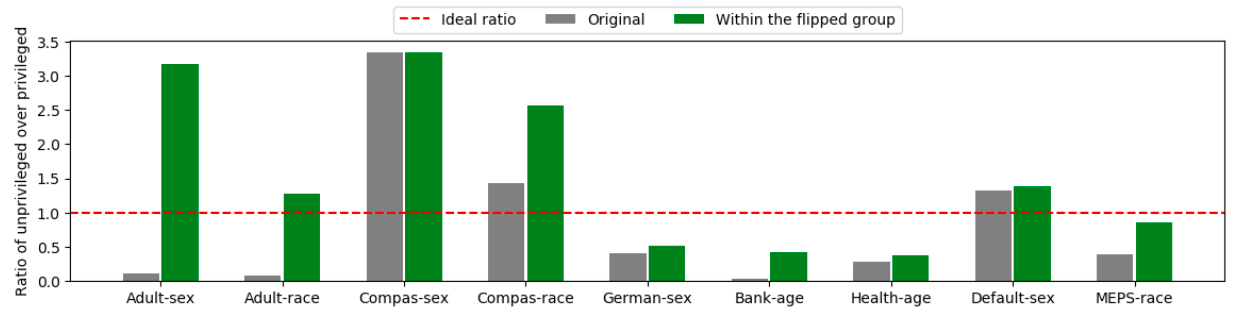}
\caption{Ratio for {\it favorable} labels. The gray bar shows the ratio of unprivileged instances receiving favorable labels among all testing data; The green bar shows the same ratio, but only among instances whose protected attribute values are flipped.} 
\label{fig:pos}
\end{figure*}
\begin{figure*}[b!]
\centering
\includegraphics[width=.90\textwidth]{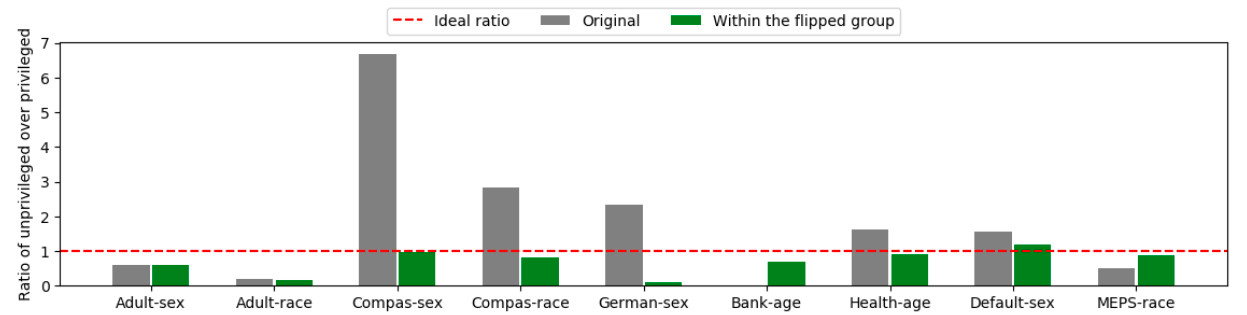}
\caption{Ratio for {\it unfavorable} labels. The gray bar shows the ratio of unprivileged instances receiving unfavorable labels among all testing data; The green bar shows the same ratio, but only among instances whose protected attribute values are mutated using our extrapolation model.} 
\label{fig:neg}
\end{figure*}
One of the most crucial presumptions in this paper (as well as the fairness domain) is that 
\begin{blockquote}
\textbf{Conjecture I}: The protected attribute itself is essentially irrelevant to the classification problem, yet it may indicate a latent correlation with other non-protected attributes.
\end{blockquote}
For example, in an ideal case, the gender of an individual should not affect the result of the loan application. Based on this presumption, we can deduce that the protected attributes in the training data of some classification problems are informative only because it is a proxy for other relevant information. 
In support of our conjecture, prior studies believe that one cause of bias is the {\it negative legacy} which means the training data previously collected is either wrongly labeled or it reflects some discriminatory practices in the past~\cite{kamishima2011fairness}. Either way, when the classification model is trained on such data and negative legacy disappears in data collected later (either because the data is correctly labeled or the discriminatory practices are eliminated), the model will generate biased outcomes that favor the privileged groups. In other words, due to data mislabeling or imbalance among groups, protected attributes can become a proxy that represents the latent correlation of other really informative (non-protected) attributes~\cite{grari2021fairness}.

Therefore, to further verify our conjecture, we investigate the negative legacy potentially embedded in the training data. To achieve this, we decided to explore whether we could "reverse engineer" the proxy, which is represented by the protected attribute. Here the base assumption is:
\bi
\item
All the informative attributes are included in the dataset;
\item
Thus, we can infer the protected attributes into some combinations of other non-protected attributes.
\ei
As shown in Figure~\ref{fig:example1}, we used linear regression (middle figure) and decision tree (right-hand-side figure) to extrapolate the correlation between the protected and non-protected attributes. In both models, we use the protected attribute (sex) as the dependent feature and other non-protected ones as independent features.
For example, as shown on the right-hand side of Figure~\ref{fig:example1}, the rule list provided in the Adult income problem reveals that the privileged group (male) is more like to possess higher capital gain and working hours within training data. Therefore, it is possible that the classification model values the protected attribute only because it has a positive correlation with "high capital balance" and "more stable job". That means the protected attribute is simply a kernel of a more profound relationship of several actually informative attributes. In that case, we could mitigate bias by decomposing the "proxy" and re-emphasizing the importance of those attributes represented by the proxy. 

Admittedly, our approach cannot guarantee the success of explaining bias (e.g., protected attributes show weak correlations with all non-protected attributes). In such a case, we believe that the protected attribute may contain some information that the dataset has not collected so far. 
That is, the trade-off between fairness and model performance can be insolvable in that scenario. 
However, within the scope of the empirical study conducted in this paper, our approach is proven to be generally effective as supported by the experiment results.

\subsection{Using Explanations for Bias Mitigation}\label{xplainbm}
 
Now as we can extrapolate the cause of bias in terms of the relationship between the protected and non-protected attributes, we seek for means to mitigate such bias.  
Our intuition is simple: If the prediction model exhibits bias that comes from data imbalance among protected attributes, we should offset such bias by relabeling the protected attributes (either assigned instances from privileged group to unprivileged group or vice versa) on certain instances.
In summary, the second conjecture in this paper is:
\begin{blockquote}
\textbf{Conjecture II}: If we can foresee which instances are more likely to be discriminated based on their protected attributes, we can try to offset such bias by masking their actual protected attributes when inferring to a biased model.
\end{blockquote}
To identify the subset of testing instances that require relabelling on protected attributes, we use the extrapolation model trained on the imbalanced training data. 
Using the Adult dataset as shown in Figure~\ref{fig:example1} as an example, we find that one of the specific causes of bias here is that the imbalanced data shows a strong correlation between the privileged group ($sex $=$ male$) and the number of working hours per week ({\it hours-per-week}), which is also positively related to the favorable class label. Now assume a new data instance in testing data possesses high {\it hours-per-week} yet an unprivileged protected attribute ($sex $=$ female$). While a high {\it hours-per-week} attribute value increases the probability of a favorable label, the unprivileged protected attribute will conversely increase the probability of receiving an unfavorable label. Thus, as we can foresee that this new instance is likely to be discriminated despite a high {\it hours-per-week} attribute, we could mask its gender into the privileged group (from female to male), which will increase the probability of favorable label.
It is noteworthy that similar conjecture was also mentioned in prior study. Zliobaite et al. introduced the concept of conditional discrimination in their paper, where they argue that if certain decision-making differences across the protected attributes are explainable, that such differences are tolerable~\cite{vzliobaite2011handling}. By distinguishing conditional discrimination, their approach can remove only the bad discrimination while allowing explainable non-discriminatory differences.  
 
To examine the applicability of our tactic, we conducted experiments that lead to preliminary results shown in Figure~\ref{fig:pos} and Figure~\ref{fig:neg}. Figure~\ref{fig:pos} plots, within testing data of each dataset, the ratio of unprivileged over privileged groups in receiving \underline{favorable} labels; Figure~\ref{fig:neg} plots the ratio of the same two groups in receiving \underline{unfavorable} labels. 
The ideal equilibrium is $ratio=1$, where privileged and unprivileged groups are evenly distributed in both classes. 
However, as revealed in Figure~\ref{fig:pos}, the unprivileged group is highly under-represented in 6 out of 9 cases. For the other 3 cases (Compas and Default datasets), Figure~\ref{fig:neg} shows that the unprivileged group suffers from an extremely higher probability of receiving unfavorable labels than the privileged group does. Fortunately, such an imbalanced ratio is diminished within the group of instances whose protected attributes are flipped by our extrapolation model (represented as green bars). 
Presented by both figures, the flipped group shows either (a) an increased ratio of an unprivileged group receiving favorable labels, or (b) a decreased ratio of an unprivileged group receiving unfavorable labels in each dataset. It is especially noteworthy that in certain datasets (Bank and MEPS) where the ratio is below 1 in both scenarios, the flipped group constantly shows a tendency of moving towards the ideal equilibrium. This indicates that our tactic is self-adaptive for efficiently handling various types of imbalance.

\subsection{FairMask Implementation}\label{FairMask}
FairMask algorithm is a hybrid method of pre-processing and post-processing. Traditional pre-processing methods are usually used to modify the training data such that a fairer model can be trained on the cleaned data. This process usually takes place before the training stage. Post-processing methods, on the other hand, are usually used after the inference stage. Given the prediction outcome from the model, a post-processing method then selectively changes the final predictions on certain instances. However, unlike either genre, FairMask is designed to be applied after the training stage\footnote{In fact, FairMask can be applied parallel to the training phase in deployment since it won't modify the training data. From our implementation, we put FairMask process after the training phase}, and before the inference stage.  
Specifically, FairMask follows the steps below:
\begin{enumerate}
\item FairMask initiates an extrapolation model which is trained on non-protected attributes and uses the protected attribute as the dependent feature. 
\item FairMask uses the extrapolation model to predict on the new incoming data (e.g., testing data).
\item Then, if the predictions disagree with the original value, we claim that the new data instance is more likely to be discriminated by the prediction model.
\item Finally, when passing this new instance to the classification model, we mask its original protected attributes by the predicted values from FairMask. 
\end{enumerate}
The first step is based on our first conjecture, that the cause of bias within a model can be explained, in forms of the latent correlation between protected and non-protected attributes. The second and third steps are based on our second conjecture. These steps aim to forecast, based on the results of explaining bias, which new instances are more likely to suffer from such bias.
The difference between the original testing data $T_{test}$ and the testing data $T_{test}'$  masked by FairMask is that $T_{test}'$ only contains synthetic protected attributes $P_{test}'$, which do not require access of the original and real protected attributes $P_{test}$. However, $P_{test}$ can still be kept for fairness evaluation purposes.

\begin{figure}[t!]
\centering
\includegraphics[width=.45\textwidth]{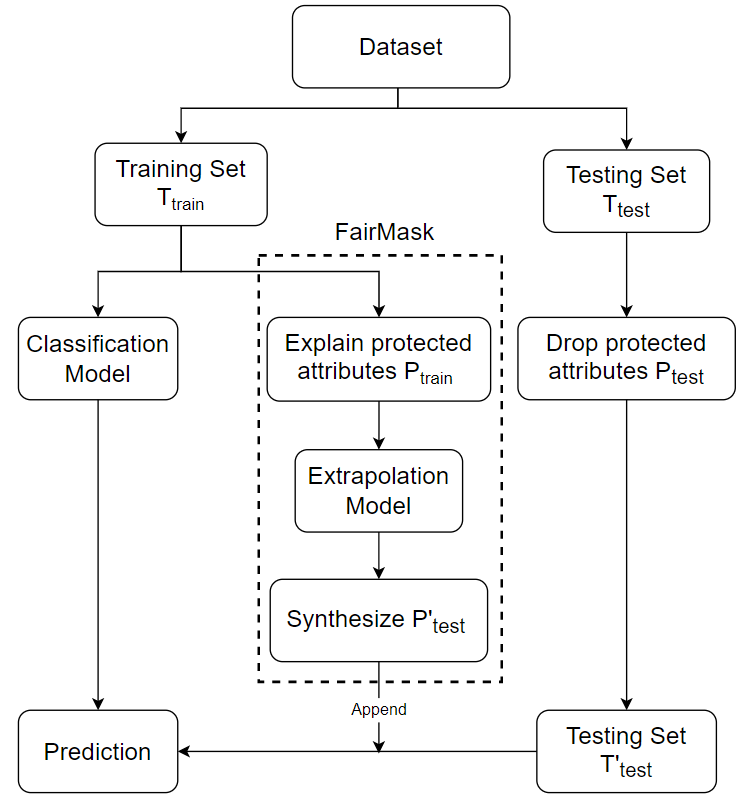}
\caption{An overview of FairMask and the experiment rig in this paper. 
Note that the synthesized protected attributes are only used for model prediction whereas the actual protected attributes are used in computing fairness metrics. 
} 
\label{fig:frame}
\end{figure}

The overview of the proposed approach is shown in Figure~\ref{fig:frame}. Algorithm~\ref{algo1} describes the pseudocode of FairMask. Please not that the actual protected attributes in testing data remain unknown to the prediction model, which as discussed later, ensures procedural justice.

The advantages of our approach over prior methods are obvious. By deploying an extrapolation model to both explain and mitigate bias, FairMask can offer concise insights on the potential cause of bias. 
As provided in Figure~\ref{fig:example1}, either linear coefficients or rule-based summaries can be used to interpret how the protected attribute ``misleads'' the classification model into algorithmic discrimination. Moreover, since FairMask does not require generating additional synthetic data samples to distort the original training samples, its runtime is much faster than the benchmark methods. Note that FairMask only uses  SMOTE~\cite{Chawla:2002} when training the extrapolation model (to better predict protected attributes), and will not affect the training data of the model.   

\begin{algorithm}[!t]
\caption{FairMask pseudocode\label{algo1}}
\small
\KwData{$T_{\mathit{train}}$ contains training data without dependent attributes;
$T_{\mathit{test}}$ contains testing data without dependent attributes;
$\mathit{budget}$ is the number of extrapolation models that can be used for weighted-vote the synthetic values
}
\KwResult{Testing data with synthetic protected attribute values $T_{test}'$}
\Begin{
$P_{\mathit{train}}$, ${NP}_{\mathit{train}}$ $\gets$ $T_{\mathit{train}}$ // Divide independent attributes into protected and non-protected attributes\\
$M$ $\gets$ InitializeModels($budget$) \\
\For{$i\gets0$ \KwTo $budget$}{
    ($P_{\mathit{train}}', NP_{\mathit{train}}'$ $\gets$ SMOTE($P_{\mathit{train}}, NP_{\mathit{train}}$) \\
    $M_{i}$ $\gets$ FitModel($P_{\mathit{train}}', NP_{\mathit{train}}'$) \\
    }
$P_{\mathit{test}}, NP_{\mathit{test}}$ $\gets$ $T_{\mathit{test}}$\\
$P_{\mathit{test}}'$ $\gets$ M.weightedVote($NP_{\mathit{test}}$)\\

$T_{\mathit{test}}'$ $\gets$ Append$(P_{\mathit{test}}',NP_{\mathit{test}})$\\


return $T_{\mathit{test}}'$}
\end{algorithm}

\section{Experiment Setup}\label{experiment}

In this section, we describe the data preparation for the experiment as well as the general experiment setup.
\subsection{Data}\label{data}
This paper uses collected datasets that are widely used in prior related research (see Table~\ref{tab:dataset}). After data collection, we first need to pre-process the data. For most of the datasets used in this paper (German, Bank, Heart, Default, and MEPS15), no feature engineering is required because either the features are all numerical or a standard procedure is adopted by all prior practitioners. For Adult and Compas datasets, there are some variants of pre-processing being proposed by past researchers. Here we did not follow the pre-processing steps mentioned in AIF360~\cite{bellamy2018ai}, which includes one-hot encoding of non-ordinal categorical features. This is because much prior work, including Fair-SMOTE~\cite{Chakraborty2021BiasIM} the benchmark method used in this paper, is only applicable to numerical and ordinal categorical features. For example, Fair-SMOTE applies deferential evolutionary algorithms to generate mutants for the purpose of over-sampling. Such methods cannot cope with the restraints from one-hot encoded features and, therefore, may generate invalid mutants. In short, we removed all non-ordinal categorical features in these two datasets. Similar approaches can also be found in many other previous works~\cite{Chakraborty2021BiasIM,feldman2015certifying,calmon2017optimized,salazar2021automated}.   

Finally, we apply min-max scaling (scale numerical values to the range of $[0,1]$ by the minimum and maximum values in each feature) to transform each dataset. For each experiment trial, we split the data into 80\% training data and 20\% testing data, using the same set of random seeds on all methods to control the variable of comparison. We repeat this procedure 20 times for statistical analysis. 

\begin{table}[t]
\centering
\caption{Performance metrics used in this paper.}
\begin{tabular}{cc}
\toprule
Metrics   & Definition                                    \\ \hline
Accuray   & (TP$+$TN)$/$(TP$+$TN$+$FP$+$FN)                         \\ 
Precision & TP$/$(TP$+$FP)                                    \\ 
Recall    & TP$/$(TP$+$FN)                                    \\ \
F1 score  & 2 $\times$ (Precision $\times$ Recall)$/$(Precision $+$ Recall) \\ 
\bottomrule
\end{tabular}
\label{tab:metrics2}
\end{table}

\begin{table*}[t!]
\centering
\scriptsize
\caption{Preliminary result on choice of the extrapolation model in FairMask. Here, cells marked in darker colors are better than those marked in lighter colors within the same dataset block. For each dataset, we repeat the experiment for 20 runs and report the median values. The ranks indicated by colors are determined by the Scott-Knott test as described in \S\ref{stats}. }
\begin{tabular}{|c|c|c|cccc|cccc|}
\hline
\begin{tabular}[c]{@{}c@{}}Dataset:\\ Protected Attribute\end{tabular} &
  \begin{tabular}[c]{@{}c@{}}Classification\\  Model\end{tabular} &
  \begin{tabular}[c]{@{}c@{}}Extrapolation\\ Model\end{tabular} &
  Accuracy &
  Precision &
  Recall &
  F1 &
  AOD &
  EOD &
  SPD &
  DI \\ \hline
 &
   &
  CART &
  \cellcolor[HTML]{C0C0C0} 83 &
  \cellcolor[HTML]{C0C0C0} 68 &
  \cellcolor[HTML]{C0C0C0} 56 &
  \cellcolor[HTML]{C0C0C0} 61 &
  \cellcolor[HTML]{C0C0C0} 2 &
  \cellcolor[HTML]{EFEFEF} 6 &
  \cellcolor[HTML]{EFEFEF} 11 &
  \cellcolor[HTML]{C0C0C0} 46 \\ \cline{3-3}
 &
  \multirow{-2}{*}{RF} &
  LR &
  \cellcolor[HTML]{C0C0C0} 83 &
  \cellcolor[HTML]{C0C0C0} 68 &
  \cellcolor[HTML]{C0C0C0} 53 &
  \cellcolor[HTML]{C0C0C0} 59 &
  \cellcolor[HTML]{C0C0C0} 2 &
  \cellcolor[HTML]{EFEFEF} 6 &
  \cellcolor[HTML]{EFEFEF} 11 &
  \cellcolor[HTML]{C0C0C0} 47 \\ \cline{2-3}
 &
   &
  CART &
  \cellcolor[HTML]{EFEFEF} 82 &
  \cellcolor[HTML]{EFEFEF} 65 &
  \cellcolor[HTML]{EFEFEF} 51 &
  \cellcolor[HTML]{EFEFEF} 57 &
  \cellcolor[HTML]{EFEFEF} 3 &
  \cellcolor[HTML]{EFEFEF} 6 &
  \cellcolor[HTML]{EFEFEF} 10 &
  \cellcolor[HTML]{EFEFEF} 48 \\ \cline{3-3}
 &
  \multirow{-2}{*}{MLP} &
  LR &
  \cellcolor[HTML]{EFEFEF} 82 &
  \cellcolor[HTML]{EFEFEF} 66 &
  \cellcolor[HTML]{EFEFEF} 48 &
  \cellcolor[HTML]{EFEFEF} 55 &
  \cellcolor[HTML]{EFEFEF} 3 &
  \cellcolor[HTML]{EFEFEF} 7 &
  \cellcolor[HTML]{EFEFEF} 11 &
  \cellcolor[HTML]{EFEFEF} 5 \\ \cline{2-3}
 &
   &
  CART &
   80 &
  \cellcolor[HTML]{EFEFEF} 67 &
   30 &
   42 &
  \cellcolor[HTML]{C0C0C0} 2 &
  \cellcolor[HTML]{C0C0C0} 1 &
  \cellcolor[HTML]{C0C0C0} 6 &
  \cellcolor[HTML]{C0C0C0} 47 \\
\multirow{-6}{*}{Adult: Sex} &
  \multirow{-2}{*}{NB} &
  LR &
   80 &
  \cellcolor[HTML]{EFEFEF} 66 &
   31 &
   42 &
  \cellcolor[HTML]{C0C0C0} 2 &
  \cellcolor[HTML]{C0C0C0} 1 &
  \cellcolor[HTML]{C0C0C0} 6 &
  \cellcolor[HTML]{C0C0C0} 47 \\ \hline
 &
   &
  CART &
  \cellcolor[HTML]{C0C0C0} 82 &
  \cellcolor[HTML]{C0C0C0} 72 &
  \cellcolor[HTML]{C0C0C0} 53 &
  \cellcolor[HTML]{C0C0C0} 61 &
  \cellcolor[HTML]{C0C0C0}0 &
  \cellcolor[HTML]{EFEFEF} 3 &
  \cellcolor[HTML]{EFEFEF} 7 &
  \cellcolor[HTML]{EFEFEF} 36 \\ \cline{3-3}
 &
  \multirow{-2}{*}{RF} &
  LR &
  \cellcolor[HTML]{C0C0C0} 83 &
  \cellcolor[HTML]{C0C0C0} 71 &
  \cellcolor[HTML]{C0C0C0} 53 &
  \cellcolor[HTML]{C0C0C0} 61 &
  \cellcolor[HTML]{C0C0C0} 1 &
  \cellcolor[HTML]{EFEFEF} 2 &
  \cellcolor[HTML]{EFEFEF} 7 &
  \cellcolor[HTML]{EFEFEF} 35 \\ \cline{2-3}
 &
   &
  CART &
  \cellcolor[HTML]{C0C0C0} 83 &
  \cellcolor[HTML]{C0C0C0} 71 &
  \cellcolor[HTML]{EFEFEF} 48 &
  \cellcolor[HTML]{EFEFEF} 57 &
  \cellcolor[HTML]{C0C0C0} 0 &
  \cellcolor[HTML]{EFEFEF} 2 &
  \cellcolor[HTML]{EFEFEF} 6 &
  \cellcolor[HTML]{EFEFEF} 37 \\ \cline{3-3}
 &
  \multirow{-2}{*}{MLP} &
  LR &
  \cellcolor[HTML]{C0C0C0} 83 &
  \cellcolor[HTML]{C0C0C0} 71 &
  \cellcolor[HTML]{EFEFEF} 49 &
  \cellcolor[HTML]{EFEFEF} 58 &
  \cellcolor[HTML]{C0C0C0} 0 &
  \cellcolor[HTML]{EFEFEF} 2 &
  \cellcolor[HTML]{EFEFEF} 6 &
  \cellcolor[HTML]{EFEFEF} 35 \\ \cline{2-3}
 &
   &
  CART &
  \cellcolor[HTML]{EFEFEF} 80 &
  \cellcolor[HTML]{EFEFEF} 67 &
   3 &
   42 &
  \cellcolor[HTML]{EFEFEF} 2 &
  \cellcolor[HTML]{C0C0C0} 1 &
  \cellcolor[HTML]{C0C0C0} 3 &
  \cellcolor[HTML]{C0C0C0} 28 \\ \cline{3-3}
\multirow{-6}{*}{Adult: Race} &
  \multirow{-2}{*}{NB} &
  LR &
  \cellcolor[HTML]{EFEFEF} 80 &
  \cellcolor[HTML]{EFEFEF} 67 &
   31 &
   42 &
  \cellcolor[HTML]{EFEFEF} 2 &
  \cellcolor[HTML]{C0C0C0} 1 &
  \cellcolor[HTML]{C0C0C0} 3 &
  \cellcolor[HTML]{C0C0C0} 3 \\ \hline
 &
   &
  CART &
  \cellcolor[HTML]{EFEFEF} 65 &
  \cellcolor[HTML]{EFEFEF} 66 &
   72 &
  \cellcolor[HTML]{EFEFEF} 69 &
  \cellcolor[HTML]{C0C0C0} 0 &
  \cellcolor[HTML]{C0C0C0} 5 &
  \cellcolor[HTML]{C0C0C0} 8 &
  \cellcolor[HTML]{C0C0C0} 12 \\ \cline{3-3}
 &
  \multirow{-2}{*}{RF} &
  LR &
  \cellcolor[HTML]{EFEFEF} 64 &
  \cellcolor[HTML]{EFEFEF} 66 &
   74 &
  \cellcolor[HTML]{EFEFEF} 7 &
  \cellcolor[HTML]{C0C0C0} 0 &
  \cellcolor[HTML]{C0C0C0} 3 &
  \cellcolor[HTML]{C0C0C0} 6 &
  \cellcolor[HTML]{C0C0C0} 1 \\ \cline{2-3}
 &
   &
  CART &
  \cellcolor[HTML]{C0C0C0} 67 &
  \cellcolor[HTML]{C0C0C0} 67 &
  \cellcolor[HTML]{EFEFEF} 79 &
  \cellcolor[HTML]{C0C0C0} 73 &
   2 &
  \cellcolor[HTML]{EFEFEF} 6 &
  \cellcolor[HTML]{EFEFEF} 11 &
  \cellcolor[HTML]{EFEFEF} 16 \\ \cline{3-3}
 &
  \multirow{-2}{*}{MLP} &
  LR &
  \cellcolor[HTML]{C0C0C0} 68 &
  \cellcolor[HTML]{C0C0C0} 68 &
  \cellcolor[HTML]{EFEFEF} 79 &
  \cellcolor[HTML]{C0C0C0} 73 &
  \cellcolor[HTML]{EFEFEF} 1 &
  \cellcolor[HTML]{EFEFEF} 6 &
  \cellcolor[HTML]{EFEFEF} 11 &
  \cellcolor[HTML]{EFEFEF} 16 \\ \cline{2-3}
 &
   &
  CART &
  \cellcolor[HTML]{C0C0C0} 67 &
  \cellcolor[HTML]{EFEFEF} 66 &
  \cellcolor[HTML]{C0C0C0} 82 &
  \cellcolor[HTML]{C0C0C0} 73 &
   3 &
   8 &
   14 &
   17 \\ \cline{3-3}
\multirow{-6}{*}{Compass: Sex} &
  \multirow{-2}{*}{NB} &
  LR &
  \cellcolor[HTML]{C0C0C0} 68 &
  \cellcolor[HTML]{C0C0C0} 67 &
  \cellcolor[HTML]{C0C0C0} 81 &
  \cellcolor[HTML]{C0C0C0} 73 &
  \cellcolor[HTML]{EFEFEF} 1 &
   8 &
   13 &
   17 \\ \hline
 &
   &
  CART &
  \cellcolor[HTML]{EFEFEF} 64 &
   66 &
   73 &
  \cellcolor[HTML]{EFEFEF} 69 &
  \cellcolor[HTML]{C0C0C0} 2 &
  \cellcolor[HTML]{C0C0C0} 9 &
  \cellcolor[HTML]{C0C0C0} 13 &
  \cellcolor[HTML]{C0C0C0} 19 \\ \cline{3-3}
 &
  \multirow{-2}{*}{RF} &
  LR &
  \cellcolor[HTML]{EFEFEF} 64 &
   66 &
   72 &
  \cellcolor[HTML]{EFEFEF} 69 &
  \cellcolor[HTML]{C0C0C0} 2 &
  \cellcolor[HTML]{C0C0C0} 9 &
  \cellcolor[HTML]{C0C0C0} 13 &
  \cellcolor[HTML]{C0C0C0} 19 \\ \cline{2-3}
 &
   &
  CART &
  \cellcolor[HTML]{C0C0C0} 69 &
  \cellcolor[HTML]{C0C0C0} 70 &
  \cellcolor[HTML]{EFEFEF} 76 &
  \cellcolor[HTML]{C0C0C0} 73 &
  \cellcolor[HTML]{EFEFEF} 3 &
  \cellcolor[HTML]{EFEFEF} 12 &
  \cellcolor[HTML]{EFEFEF} 18 &
  \cellcolor[HTML]{EFEFEF} 25 \\ \cline{3-3}
 &
  \multirow{-2}{*}{MLP} &
  LR &
  \cellcolor[HTML]{C0C0C0} 69 &
  \cellcolor[HTML]{C0C0C0} 70 &
  \cellcolor[HTML]{EFEFEF} 76 &
  \cellcolor[HTML]{C0C0C0} 73 &
  \cellcolor[HTML]{EFEFEF} 3 &
  \cellcolor[HTML]{EFEFEF} 12 &
  \cellcolor[HTML]{EFEFEF} 18 &
  \cellcolor[HTML]{EFEFEF} 25 \\ \cline{2-3}
 &
   &
  CART &
  \cellcolor[HTML]{C0C0C0} 68 &
  \cellcolor[HTML]{EFEFEF} 68 &
  \cellcolor[HTML]{C0C0C0} 79 &
  \cellcolor[HTML]{C0C0C0} 73 &
  \cellcolor[HTML]{EFEFEF} 4 &
  \cellcolor[HTML]{EFEFEF} 11 &
  \cellcolor[HTML]{EFEFEF} 18 &
  \cellcolor[HTML]{EFEFEF} 24 \\ \cline{3-3}
\multirow{-6}{*}{Compass: Race} &
  \multirow{-2}{*}{NB} &
  LR &
  \cellcolor[HTML]{C0C0C0} 68 &
  \cellcolor[HTML]{EFEFEF} 68 &
  \cellcolor[HTML]{C0C0C0} 78 &
  \cellcolor[HTML]{C0C0C0} 73 &
  \cellcolor[HTML]{EFEFEF} 4 &
  \cellcolor[HTML]{EFEFEF} 11 &
  \cellcolor[HTML]{EFEFEF} 18 &
  \cellcolor[HTML]{EFEFEF} 24 \\ \hline
 &
  &
  CART &
  \cellcolor[HTML]{C0C0C0} 70 &
  \cellcolor[HTML]{EFEFEF} 73 &
  \cellcolor[HTML]{C0C0C0} 92 &
  \cellcolor[HTML]{C0C0C0} 81 &
  \cellcolor[HTML]{EFEFEF} 5 &
  \cellcolor[HTML]{C0C0C0} 4 &
  \cellcolor[HTML]{C0C0C0} 8 &
  \cellcolor[HTML]{C0C0C0} 9 \\ \cline{3-3}
 &
  \multirow{-2}{*}{RF} &
  LR &
  \cellcolor[HTML]{C0C0C0} 69 &
  \cellcolor[HTML]{EFEFEF} 71 &
  \cellcolor[HTML]{C0C0C0} 92 &
  \cellcolor[HTML]{C0C0C0} 80 &
  \cellcolor[HTML]{EFEFEF} 5 &
  \cellcolor[HTML]{C0C0C0} 6 &
  \cellcolor[HTML]{C0C0C0} 9 &
  \cellcolor[HTML]{C0C0C0} 10 \\ \cline{2-3}
 &
  &
  CART &
  \cellcolor[HTML]{C0C0C0} 69 &
  \cellcolor[HTML]{EFEFEF} 71 &
  \cellcolor[HTML]{C0C0C0} 93 &
  \cellcolor[HTML]{C0C0C0} 81 &
  \cellcolor[HTML]{EFEFEF} 6 &
  \cellcolor[HTML]{C0C0C0} 4 &
  \cellcolor[HTML]{C0C0C0} 8 &
  \cellcolor[HTML]{C0C0C0} 9 \\ \cline{3-3}
 &
  \multirow{-2}{*}{MLP} &
  LR &
  \cellcolor[HTML]{C0C0C0} 69 &
  \cellcolor[HTML]{EFEFEF} 71 &
  \cellcolor[HTML]{C0C0C0} 92 &
  \cellcolor[HTML]{C0C0C0} 80 &
  \cellcolor[HTML]{EFEFEF} 6 &
  \cellcolor[HTML]{C0C0C0} 4 &
  \cellcolor[HTML]{C0C0C0} 9 &
  \cellcolor[HTML]{C0C0C0} 9 \\ \cline{2-3}
 &
  &
  CART &
  \cellcolor[HTML]{EFEFEF} 60 &
  \cellcolor[HTML]{C0C0C0} 79 &
  \cellcolor[HTML]{EFEFEF} 59 &
  \cellcolor[HTML]{EFEFEF} 67 &
  \cellcolor[HTML]{C0C0C0} 2 &
  \cellcolor[HTML]{EFEFEF} 11 &
  \cellcolor[HTML]{EFEFEF} 11 &
  \cellcolor[HTML]{EFEFEF} 18 \\ \cline{3-3}
\multirow{-6}{*}{German: Sex} &
  \multirow{-2}{*}{NB} &
  LR &
  \cellcolor[HTML]{EFEFEF} 60 &
  \cellcolor[HTML]{C0C0C0} 79 &
  \cellcolor[HTML]{EFEFEF} 59 &
  \cellcolor[HTML]{EFEFEF} 67 &
  \cellcolor[HTML]{C0C0C0} 2 &
  \cellcolor[HTML]{EFEFEF} 11 &
  \cellcolor[HTML]{EFEFEF} 11 &
  \cellcolor[HTML]{EFEFEF} 18 \\ \hline
\end{tabular}
\label{tab:result-model}
\end{table*}

\subsection{Model Selection}
In FairMask. we must select a {\bf classification} model and an {\bf extrapolation} model: The classification model is used to predict the dependent variable in the task using independent variables. The extrapolation model is used by FairMask to explain and mitigate the bias of the classification model.
In Table~\ref{tab:result-model}, we explore the interplay of different classification models and extrapolation models in three of the datasets used in this paper. 
Our initial choices of classification models include random forest (RF), 2-layer neural network (as known as multi-layer perceptron, MLP), and naive Bayes (NB). As for the extrapolation model, we include two highly interpretable models: logistic regression (LR) and classification and regression tree (CART). Indicated by the result, we cannot find an absolute winner among the classification models, which can outperform others in all cases. Moreover, the choice of the extrapolation model has a trivial influence on the final result. In short, our general insight from Table~\ref{tab:result-model} is that (a) the choice of the classification model varies among different datasets, and (b) the performance of FairMask is robust regardless of the choice of extrapolation model. Hence, in the following experiment, we will use RF as the classification model and CART as the extrapolation model. 

\subsection{Evaluation Criteria}
To evaluate the predictive performance of each method, we use metrics computed by the confusion matrix of binary classification: accuracy, precision, recall, and F1 score. These criteria
are selected since (a)~they are widely used in both software analytics~\cite{local_tse,menzies2014sharing} and fairness literature~\cite{9286091,Biswas_2020,hardt2016equality,pleiss2017fairness,zhang2020white}
and (b)~the latest benchmark method in our experiment, Fair-SMOTE, uses the exact same set of criteria.
The  definitions of performance metrics are shown in Table~\ref{tab:metrics2}.

To assess the effectiveness of mitigating bias, we use fairness metrics as introduced in Table~\ref{tab:metrics}, some of which are also computed based on the confusion matrix of binary classification. The group fairness metrics aim to evaluate whether different social groups, as identified by their protected attributes, receive statistically similar prediction outcomes by the classification model. The individual fairness metric, denoted as Flip Rate, is designed based on the intuition of procedural justice. By definition, when individuals that are similar to each other regardless the protected attributes, they shall receive similar prediction outcomes (in this case of binary classification, the same outcome). To assess this criterion, we use the following situation testing tactic:
\begin{itemize}
    \item For each instance in testing data, flip the protected attribute.
    \item Pass the edited data instances into the classification model
    \item Record the times where the new prediction outcome differs from the original one.
\end{itemize}
It is noteworthy that the situation testing is also used in prior work~\cite{Chakraborty2021BiasIM}. The major difference is that Fair-SMOTE uses situation testing as a technique to identify and remove bias-introducing instances whereas in this paper we only use it to assess the extent of individual fairness for all the methods examined in the experiment.

\subsection{Statistical Analysis}\label{stats}

To compare the predictive performance and ability in mitigating bias among all algorithms on every dataset, we use a non-parametric significance test along with a non-parametric effect size test. 
Specifically, we use the Scott-Knott test~\cite{mittas2012ranking} that sorts the list of treatments (in this case, the benchmark bias-mitigation methods and our approach) by their median scores. After the sorting, it then splits the list into two sub-lists. The objective for such a split is to maximize the expected value of differences $E(\Delta)$ in the observed performances before and after division~\cite{xia2018hyperparameter}:
\begin{equation}
    E(\Delta) = \frac{|l_1|}{|l|}abs(E({l_1}) - E({l}))^2 + \frac{|l_2|}{|l|}abs(E({l_2}) - E({l}))^2
\end{equation}
where $|l_1|$ means the size of list $l_1$.

The Scott-Knott test assigns ranks to each result set; the higher the rank, the better the result. Two results will be ranked the same if the difference between the distributions is not significant.
In this expression, Cliff's Delta estimates the probability that a value in list $A$ is greater than a value in list $B$, minus the reverse probability~\cite{macbeth2011cliff}.   A division passes this hypothesis test if it is not a ``small'' effect ($Delta \geq 0.147$). 
This hypothesis test and its effect sizes are supported by Hess and Kromery~\cite{hess2004robust}.

\begin{table*} 
\centering
\caption{Results for RQ2. RF denotes the default random forest learner. For all performance metrics, greater is better; for all fairness metrics, smaller is better. Here, cells marked in darker colors are better than those marked in lighter colors within the same dataset block. 
For each dataset, we repeat the experiment for 20 runs and report the median values in percentage. The ranks indicated by colors are determined by the Scott-Knott test as described in \S\ref{stats}.
}
\scriptsize 
\begin{tabular}{|
>{\columncolor[HTML]{FFFFFF}}c |
>{\columncolor[HTML]{FFFFFF}}c |
>{\columncolor[HTML]{9B9B9B}}c 
>{\columncolor[HTML]{9B9B9B}}c c
>{\columncolor[HTML]{9B9B9B}}c |ccccc|}
\hline
{\color[HTML]{333333} Dataset: Protected Attribute} &
  {\color[HTML]{333333} Method} &
  \cellcolor[HTML]{FFFFFF}{\color[HTML]{333333} Accuracy} &
  \cellcolor[HTML]{FFFFFF}{\color[HTML]{333333} Precision} &
  \cellcolor[HTML]{FFFFFF}{\color[HTML]{333333} Recall} &
  \cellcolor[HTML]{FFFFFF}{\color[HTML]{333333} F1 score} &
  \cellcolor[HTML]{FFFFFF}{\color[HTML]{333333} AOD} &
  \cellcolor[HTML]{FFFFFF}{\color[HTML]{333333} EOD} &
  \cellcolor[HTML]{FFFFFF}{\color[HTML]{333333} SPD} &
  \cellcolor[HTML]{FFFFFF}{\color[HTML]{333333} DI} &
  \cellcolor[HTML]{FFFFFF}{\color[HTML]{333333} FR} \\ \hline
\cellcolor[HTML]{FFFFFF}{\color[HTML]{333333} } &
  {\color[HTML]{333333} RF} &
  {\color[HTML]{333333} { 83} } &
  {\color[HTML]{333333} { 72}} &
  \cellcolor[HTML]{C0C0C0}{\color[HTML]{333333}  53} &
  {\color[HTML]{333333}   { 61}} &
  \cellcolor[HTML]{EFEFEF}{\color[HTML]{333333}   8} &
  \cellcolor[HTML]{EFEFEF}{\color[HTML]{333333}  24} &
  \cellcolor[HTML]{EFEFEF}{\color[HTML]{333333}  18} &
  \cellcolor[HTML]{EFEFEF}{\color[HTML]{333333}  78} &
  \cellcolor[HTML]{EFEFEF}{\color[HTML]{333333}  20} \\ \cline{2-2}
\cellcolor[HTML]{FFFFFF}{\color[HTML]{333333} } &
  {\color[HTML]{333333} RF+Random} &
  {\color[HTML]{333333}   { 83}} &
  {\color[HTML]{333333}   { 75}} &
  \cellcolor[HTML]{EFEFEF}{\color[HTML]{333333}  48} &
  \cellcolor[HTML]{C0C0C0}{\color[HTML]{333333}  57} &
  \cellcolor[HTML]{9B9B9B}{\color[HTML]{333333}   {  1}} &
  \cellcolor[HTML]{9B9B9B}{\color[HTML]{333333}   {  4}} &
  \cellcolor[HTML]{9B9B9B}{\color[HTML]{333333}   { 11}} &
  \cellcolor[HTML]{C0C0C0}{\color[HTML]{333333}  52} &
  \cellcolor[HTML]{C0C0C0}{\color[HTML]{333333}   7} \\ \cline{2-2}
\cellcolor[HTML]{FFFFFF}{\color[HTML]{333333} } &
  {\color[HTML]{333333} RF+Reweighing} &
  \cellcolor[HTML]{EFEFEF}{\color[HTML]{333333}  75} &
  \cellcolor[HTML]{EFEFEF}{\color[HTML]{333333}  48} &
  \cellcolor[HTML]{9B9B9B}{\color[HTML]{333333}   { 71}} &
  \cellcolor[HTML]{C0C0C0}{\color[HTML]{333333}  57} &
  \cellcolor[HTML]{9B9B9B}{\color[HTML]{333333}   {  1}} &
  \cellcolor[HTML]{9B9B9B}{\color[HTML]{333333}   {  7}} &
  \cellcolor[HTML]{C0C0C0}{\color[HTML]{333333}  15} &
  \cellcolor[HTML]{9B9B9B}{\color[HTML]{333333}   { 37}} &
  \cellcolor[HTML]{9B9B9B}{\color[HTML]{333333}   {  2}} \\ \cline{2-2}
\cellcolor[HTML]{FFFFFF}{\color[HTML]{333333} } &
  {\color[HTML]{333333} RF+Fair-SMOTE} &
  \cellcolor[HTML]{C0C0C0}{\color[HTML]{333333}  79} &
  \cellcolor[HTML]{FFFFFF}{\color[HTML]{333333}  54} &
  \cellcolor[HTML]{9B9B9B}{\color[HTML]{333333}   { 71}} &
  {\color[HTML]{333333}   { 61}} &
  \cellcolor[HTML]{EFEFEF}{\color[HTML]{333333}   6} &
  \cellcolor[HTML]{EFEFEF}{\color[HTML]{333333}  22} &
  \cellcolor[HTML]{EFEFEF}{\color[HTML]{333333}  20} &
  \cellcolor[HTML]{C0C0C0}{\color[HTML]{333333}  54} &
  \cellcolor[HTML]{EFEFEF}{\color[HTML]{333333}  18} \\ \cline{2-2}
\cellcolor[HTML]{FFFFFF}{\color[HTML]{333333} } &
  {\color[HTML]{333333} RF+ROC} &
  \cellcolor[HTML]{EFEFEF}{\color[HTML]{333333}  75} &
  \cellcolor[HTML]{EFEFEF}{\color[HTML]{333333}  49} &
  \cellcolor[HTML]{9B9B9B}{\color[HTML]{333333}   {76}} &
  {\color[HTML]{333333}   { 60}} &
  \cellcolor[HTML]{C0C0C0}{\color[HTML]{333333}   2} &
  \cellcolor[HTML]{C0C0C0}{\color[HTML]{333333}  11} &
  \cellcolor[HTML]{EFEFEF}{\color[HTML]{333333}  18} &
  \cellcolor[HTML]{9B9B9B}{\color[HTML]{333333}  40} &
  \cellcolor[HTML]{EFEFEF}{\color[HTML]{333333}  24} \\ \cline{2-2}
\multirow{-5}{*}{\cellcolor[HTML]{FFFFFF}{\color[HTML]{333333} Adult: Sex}} &
  {\color[HTML]{333333} RF+FairMask} &
  {\color[HTML]{333333}   { 83}} &
  \cellcolor[HTML]{C0C0C0}{\color[HTML]{333333}  67} &
  \cellcolor[HTML]{C0C0C0}{\color[HTML]{333333}  56} &
  {\color[HTML]{333333}   { 61}} &
  \cellcolor[HTML]{9B9B9B}{\color[HTML]{333333}   {  2}} &
  \cellcolor[HTML]{9B9B9B}{\color[HTML]{333333}   {  6}} &
  \cellcolor[HTML]{9B9B9B}{\color[HTML]{333333}   { 10}} &
  \cellcolor[HTML]{9B9B9B}{\color[HTML]{333333}   { 46}} &
  \cellcolor[HTML]{9B9B9B}{\color[HTML]{333333}   {0}} \\ \hline
\cellcolor[HTML]{FFFFFF}{\color[HTML]{333333} } &
  {\color[HTML]{333333} RF} &
  {\color[HTML]{333333}  84} &
  {\color[HTML]{333333}  73} &
  \cellcolor[HTML]{C0C0C0}{\color[HTML]{333333}  53} &
  {\color[HTML]{333333}  61} &
  \cellcolor[HTML]{C0C0C0}{\color[HTML]{333333}   3} &
  \cellcolor[HTML]{FFFFFF}{\color[HTML]{333333}  10} &
  \cellcolor[HTML]{EFEFEF}{\color[HTML]{333333}   9} &
  \cellcolor[HTML]{FFFFFF}{\color[HTML]{333333}  49} &
  \cellcolor[HTML]{EFEFEF}{\color[HTML]{333333}   9} \\ \cline{2-2}
\cellcolor[HTML]{FFFFFF}{\color[HTML]{333333} } &
  {\color[HTML]{333333} RF+Random} &
  {\color[HTML]{333333}  83} &
  {\color[HTML]{333333}  70} &
  \cellcolor[HTML]{EFEFEF}{\color[HTML]{333333}  51} &
  \cellcolor[HTML]{C0C0C0}{\color[HTML]{333333}  59} &
  \cellcolor[HTML]{9B9B9B}{\color[HTML]{333333} 0} &
  \cellcolor[HTML]{9B9B9B}{\color[HTML]{333333}   3} &
  \cellcolor[HTML]{C0C0C0}{\color[HTML]{333333}   8} &
  \cellcolor[HTML]{EFEFEF}{\color[HTML]{333333}  43} &
  \cellcolor[HTML]{EFEFEF}{\color[HTML]{333333}   8} \\ \cline{2-2}
\cellcolor[HTML]{FFFFFF}{\color[HTML]{333333} } &
  {\color[HTML]{333333} RF+Reweighing} &
  \cellcolor[HTML]{C0C0C0}{\color[HTML]{333333}  76} &
  \cellcolor[HTML]{EFEFEF}{\color[HTML]{333333}  49} &
  \cellcolor[HTML]{9B9B9B}{\color[HTML]{333333}  72} &
  \cellcolor[HTML]{C0C0C0}{\color[HTML]{333333}  59} &
  \cellcolor[HTML]{C0C0C0}{\color[HTML]{333333}   3} &
  \cellcolor[HTML]{C0C0C0}{\color[HTML]{333333}   5} &
  \cellcolor[HTML]{9B9B9B}{\color[HTML]{333333}   5} &
  \cellcolor[HTML]{9B9B9B}{\color[HTML]{333333}  12} &
  \cellcolor[HTML]{C0C0C0}{\color[HTML]{333333}   4} \\ \cline{2-2}
\cellcolor[HTML]{FFFFFF}{\color[HTML]{333333} } &
  {\color[HTML]{333333} RF+Fair-SMOTE} &
  \cellcolor[HTML]{C0C0C0}{\color[HTML]{333333}  79} &
  \cellcolor[HTML]{C0C0C0}{\color[HTML]{333333}  54} &
  \cellcolor[HTML]{9B9B9B}{\color[HTML]{333333}  72} &
  {\color[HTML]{333333}  62} &
  \cellcolor[HTML]{C0C0C0}{\color[HTML]{333333}   2} &
  \cellcolor[HTML]{EFEFEF}{\color[HTML]{333333}   7} &
  \cellcolor[HTML]{EFEFEF}{\color[HTML]{333333}  11} &
  \cellcolor[HTML]{C0C0C0}{\color[HTML]{333333}  37} &
  \cellcolor[HTML]{FFFFFF}{\color[HTML]{333333}  16} \\ \cline{2-2}
\cellcolor[HTML]{FFFFFF}{\color[HTML]{333333} } &
  {\color[HTML]{333333} RF+ROC} &
  \cellcolor[HTML]{EFEFEF}{\color[HTML]{333333}  73} &
  \cellcolor[HTML]{EFEFEF}{\color[HTML]{333333}  45} &
  \cellcolor[HTML]{9B9B9B}{\color[HTML]{333333}  82} &
  \cellcolor[HTML]{C0C0C0}{\color[HTML]{333333}  59} &
  \cellcolor[HTML]{C0C0C0}{\color[HTML]{333333}   2} &
  \cellcolor[HTML]{C0C0C0}{\color[HTML]{333333}  5} &
  \cellcolor[HTML]{EFEFEF}{\color[HTML]{333333}  11} &
  \cellcolor[HTML]{9B9B9B}{\color[HTML]{333333}  25} &
  \cellcolor[HTML]{FFFFFF}{\color[HTML]{333333}  28} \\ \cline{2-2}
\multirow{-5}{*}{\cellcolor[HTML]{FFFFFF}{\color[HTML]{333333} Adult: Race}} &
  {\color[HTML]{333333} RF+FairMask} &
  {\color[HTML]{333333}  82} &
  {\color[HTML]{333333}  72} &
  \cellcolor[HTML]{C0C0C0}{\color[HTML]{333333}  53} &
  {\color[HTML]{333333}  61} &
  \cellcolor[HTML]{9B9B9B}{\color[HTML]{333333} 0} &
  \cellcolor[HTML]{9B9B9B}{\color[HTML]{333333}   3} &
  \cellcolor[HTML]{C0C0C0}{\color[HTML]{333333}   7} &
  \cellcolor[HTML]{C0C0C0}{\color[HTML]{333333}  36} &
  \cellcolor[HTML]{9B9B9B}{\color[HTML]{333333} 0} \\ \hline
\cellcolor[HTML]{FFFFFF}{\color[HTML]{333333} } &
  {\color[HTML]{333333} RF} &
  {\color[HTML]{333333}  65} &
  {\color[HTML]{333333}  67} &
  \cellcolor[HTML]{9B9B9B}{\color[HTML]{333333}  73} &
  {\color[HTML]{333333}  70} &
  \cellcolor[HTML]{EFEFEF}{\color[HTML]{333333}   5} &
  \cellcolor[HTML]{EFEFEF}{\color[HTML]{333333}  10} &
  \cellcolor[HTML]{EFEFEF}{\color[HTML]{333333}  14} &
  \cellcolor[HTML]{EFEFEF}{\color[HTML]{333333}  19} &
  \cellcolor[HTML]{EFEFEF}{\color[HTML]{333333}  28} \\ \cline{2-2}
\cellcolor[HTML]{FFFFFF}{\color[HTML]{333333} } &
  {\color[HTML]{333333} RF+Random} &
  {\color[HTML]{333333}  64} &
  {\color[HTML]{333333}  66} &
  \cellcolor[HTML]{9B9B9B}{\color[HTML]{333333}  71} &
  {\color[HTML]{333333}  68} &
  \cellcolor[HTML]{9B9B9B}{\color[HTML]{333333} 0} &
  \cellcolor[HTML]{C0C0C0}{\color[HTML]{333333}   8} &
  \cellcolor[HTML]{EFEFEF}{\color[HTML]{333333}  11} &
  \cellcolor[HTML]{EFEFEF}{\color[HTML]{333333}  16} &
  \cellcolor[HTML]{EFEFEF}{\color[HTML]{333333}  27} \\ \cline{2-2}
\cellcolor[HTML]{FFFFFF}{\color[HTML]{333333} } &
  {\color[HTML]{333333} RF+Reweighing} &
  \cellcolor[HTML]{C0C0C0}{\color[HTML]{333333}  62} &
  \cellcolor[HTML]{C0C0C0}{\color[HTML]{333333}  64} &
  \cellcolor[HTML]{C0C0C0}{\color[HTML]{333333}  67} &
  \cellcolor[HTML]{C0C0C0}{\color[HTML]{333333}  66} &
  \cellcolor[HTML]{C0C0C0}{\color[HTML]{333333}   3} &
  \cellcolor[HTML]{C0C0C0}{\color[HTML]{333333}   7} &
  \cellcolor[HTML]{EFEFEF}{\color[HTML]{333333}  12} &
  \cellcolor[HTML]{EFEFEF}{\color[HTML]{333333}  18} &
  \cellcolor[HTML]{FFFFFF}{\color[HTML]{333333}  30} \\ \cline{2-2}
\cellcolor[HTML]{FFFFFF}{\color[HTML]{333333} } &
  {\color[HTML]{333333} RF+Fair-SMOTE} &
  {\color[HTML]{333333}  65} &
  {\color[HTML]{333333}  67} &
  \cellcolor[HTML]{9B9B9B}{\color[HTML]{333333}  70} &
  {\color[HTML]{333333}  68} &
  \cellcolor[HTML]{9B9B9B}{\color[HTML]{333333} 0} &
  \cellcolor[HTML]{9B9B9B}{\color[HTML]{333333}   6} &
  \cellcolor[HTML]{C0C0C0}{\color[HTML]{333333}   9} &
  \cellcolor[HTML]{EFEFEF}{\color[HTML]{333333}  17} &
  \cellcolor[HTML]{C0C0C0}{\color[HTML]{333333}  21} \\ \cline{2-2}
\cellcolor[HTML]{FFFFFF}{\color[HTML]{333333} } &
  {\color[HTML]{333333} RF+ROC} &
  \cellcolor[HTML]{9B9B9B}{\color[HTML]{333333}  65} &
  \cellcolor[HTML]{9B9B9B}{\color[HTML]{333333}  66} &
  \cellcolor[HTML]{9B9B9B}{\color[HTML]{333333}  74} &
  \cellcolor[HTML]{9B9B9B}{\color[HTML]{333333}  70} &
  \cellcolor[HTML]{C0C0C0}{\color[HTML]{333333}   1} &
  \cellcolor[HTML]{9B9B9B}{\color[HTML]{333333}   6} &
  \cellcolor[HTML]{9B9B9B}{\color[HTML]{333333}   3} &
  \cellcolor[HTML]{9B9B9B}{\color[HTML]{333333}   4} &
  \cellcolor[HTML]{FFFFFF}{\color[HTML]{333333}  36} \\ \cline{2-2}
\multirow{-5}{*}{\cellcolor[HTML]{FFFFFF}{\color[HTML]{333333} Compas: Sex}} &
  {\color[HTML]{333333} RF+FairMask} &
  {\color[HTML]{333333}  65} &
  {\color[HTML]{333333}  66} &
  \cellcolor[HTML]{9B9B9B}{\color[HTML]{333333}  72} &
  {\color[HTML]{333333}  69} &
  \cellcolor[HTML]{9B9B9B}{\color[HTML]{333333} 0} &
  \cellcolor[HTML]{9B9B9B}{\color[HTML]{333333}   6} &
  \cellcolor[HTML]{C0C0C0}{\color[HTML]{333333}   8} &
  \cellcolor[HTML]{C0C0C0}{\color[HTML]{333333}  12} &
  \cellcolor[HTML]{9B9B9B}{\color[HTML]{333333} 0} \\ \hline
\cellcolor[HTML]{FFFFFF}{\color[HTML]{333333} } &
  {\color[HTML]{333333} RF} &
  \cellcolor[HTML]{C0C0C0}{\color[HTML]{333333}  65} &
  \cellcolor[HTML]{C0C0C0}{\color[HTML]{333333}  67} &
  \cellcolor[HTML]{9B9B9B}{\color[HTML]{333333}  73} &
  {\color[HTML]{333333}  70} &
  \cellcolor[HTML]{C0C0C0}{\color[HTML]{333333}   2} &
  \cellcolor[HTML]{EFEFEF}{\color[HTML]{333333}  10} &
  \cellcolor[HTML]{EFEFEF}{\color[HTML]{333333}  14} &
  \cellcolor[HTML]{EFEFEF}{\color[HTML]{333333}  20} &
  \cellcolor[HTML]{EFEFEF}{\color[HTML]{333333}  24} \\ \cline{2-2}
\cellcolor[HTML]{FFFFFF}{\color[HTML]{333333} } &
  {\color[HTML]{333333} RF+Random} &
  \cellcolor[HTML]{C0C0C0}{\color[HTML]{333333}  64} &
  \cellcolor[HTML]{C0C0C0}{\color[HTML]{333333}  66} &
  \cellcolor[HTML]{9B9B9B}{\color[HTML]{333333}  73} &
  {\color[HTML]{333333}  69} &
  \cellcolor[HTML]{C0C0C0}{\color[HTML]{333333}   2} &
  \cellcolor[HTML]{EFEFEF}{\color[HTML]{333333}  10} &
  \cellcolor[HTML]{EFEFEF}{\color[HTML]{333333}  14} &
  \cellcolor[HTML]{EFEFEF}{\color[HTML]{333333}  20} &
  \cellcolor[HTML]{EFEFEF}{\color[HTML]{333333}  26} \\ \cline{2-2}
\cellcolor[HTML]{FFFFFF}{\color[HTML]{333333} } &
  {\color[HTML]{333333} RF+Reweighing} &
  \cellcolor[HTML]{C0C0C0}{\color[HTML]{333333}  63} &
  \cellcolor[HTML]{C0C0C0}{\color[HTML]{333333}  66} &
  \cellcolor[HTML]{EFEFEF}{\color[HTML]{333333}  66} &
  \cellcolor[HTML]{C0C0C0}{\color[HTML]{333333}  66} &
  \cellcolor[HTML]{9B9B9B}{\color[HTML]{333333}   1} &
  \cellcolor[HTML]{9B9B9B}{\color[HTML]{333333}   2} &
  \cellcolor[HTML]{9B9B9B}{\color[HTML]{333333}   5} &
  \cellcolor[HTML]{9B9B9B}{\color[HTML]{333333}   9} &
  \cellcolor[HTML]{C0C0C0}{\color[HTML]{333333}  19} \\ \cline{2-2}
\cellcolor[HTML]{FFFFFF}{\color[HTML]{333333} } &
  {\color[HTML]{333333} RF+Fair-SMOTE} &
  \cellcolor[HTML]{C0C0C0}{\color[HTML]{333333}  65} &
  {\color[HTML]{333333}  68} &
  \cellcolor[HTML]{C0C0C0}{\color[HTML]{333333}  70} &
  {\color[HTML]{333333}  69} &
  \cellcolor[HTML]{C0C0C0}{\color[HTML]{333333}   3} &
  \cellcolor[HTML]{C0C0C0}{\color[HTML]{333333}   4} &
  \cellcolor[HTML]{C0C0C0}{\color[HTML]{333333}  13} &
  \cellcolor[HTML]{C0C0C0}{\color[HTML]{333333}  15} &
  \cellcolor[HTML]{C0C0C0}{\color[HTML]{333333}  18} \\ \cline{2-2}
\cellcolor[HTML]{FFFFFF}{\color[HTML]{333333} } &
  {\color[HTML]{333333} RF+ROC} &
  \cellcolor[HTML]{9B9B9B}{\color[HTML]{333333}  79} &
  \cellcolor[HTML]{EFEFEF}{\color[HTML]{333333}  54} &
  \cellcolor[HTML]{9B9B9B}{\color[HTML]{333333}  72} &
  \cellcolor[HTML]{9B9B9B}{\color[HTML]{333333}  68} &
  \cellcolor[HTML]{9B9B9B}{\color[HTML]{333333}   1} &
  \cellcolor[HTML]{C0C0C0}{\color[HTML]{333333}   5} &
  \cellcolor[HTML]{9B9B9B}{\color[HTML]{333333}   7} &
  \cellcolor[HTML]{9B9B9B}{\color[HTML]{333333}   12} &
  \cellcolor[HTML]{FFFFFF}{\color[HTML]{333333}  34} \\ \cline{2-2}
\multirow{-5}{*}{\cellcolor[HTML]{FFFFFF}{\color[HTML]{333333} Compas: Race}} &
  {\color[HTML]{333333} RF+FairMask} &
  \cellcolor[HTML]{C0C0C0}{\color[HTML]{333333}  64} &
  \cellcolor[HTML]{C0C0C0}{\color[HTML]{333333}  67} &
  \cellcolor[HTML]{9B9B9B}{\color[HTML]{333333}  73} &
  {\color[HTML]{333333}  69} &
  \cellcolor[HTML]{C0C0C0}{\color[HTML]{333333}   2} &
  \cellcolor[HTML]{C0C0C0}{\color[HTML]{333333}   8} &
  \cellcolor[HTML]{C0C0C0}{\color[HTML]{333333}  13} &
  \cellcolor[HTML]{EFEFEF}{\color[HTML]{333333}  19} &
  \cellcolor[HTML]{9B9B9B}{\color[HTML]{333333} 0} \\ \hline
\cellcolor[HTML]{FFFFFF}{\color[HTML]{333333} } &
  {\color[HTML]{333333} RF} &
  {\color[HTML]{333333}  70} &
  \cellcolor[HTML]{C0C0C0}{\color[HTML]{333333}  72} &
  \cellcolor[HTML]{9B9B9B}{\color[HTML]{333333}  93} &
  {\color[HTML]{333333}  81} &
  \cellcolor[HTML]{C0C0C0}{\color[HTML]{333333}   5} &
  \cellcolor[HTML]{EFEFEF}{\color[HTML]{333333}   7} &
  \cellcolor[HTML]{C0C0C0}{\color[HTML]{333333}  11} &
  \cellcolor[HTML]{C0C0C0}{\color[HTML]{333333}  11} &
  \cellcolor[HTML]{EFEFEF}{\color[HTML]{333333}  14} \\ \cline{2-2}
\cellcolor[HTML]{FFFFFF}{\color[HTML]{333333} } &
  {\color[HTML]{333333} RF+Random} &
  \cellcolor[HTML]{C0C0C0}{\color[HTML]{333333}  67} &
  \cellcolor[HTML]{C0C0C0}{\color[HTML]{333333}  71} &
  \cellcolor[HTML]{C0C0C0}{\color[HTML]{333333}  90} &
  \cellcolor[HTML]{C0C0C0}{\color[HTML]{333333}  79} &
  \cellcolor[HTML]{9B9B9B}{\color[HTML]{333333}   2} &
  \cellcolor[HTML]{9B9B9B}{\color[HTML]{333333}   1} &
  \cellcolor[HTML]{9B9B9B}{\color[HTML]{333333}   8} &
  \cellcolor[HTML]{9B9B9B}{\color[HTML]{333333}   9} &
  \cellcolor[HTML]{EFEFEF}{\color[HTML]{333333}  13} \\ \cline{2-2}
\cellcolor[HTML]{FFFFFF}{\color[HTML]{333333} } &
  {\color[HTML]{333333} RF+Reweighing} &
  \cellcolor[HTML]{EFEFEF}{\color[HTML]{333333}  64} &
  {\color[HTML]{333333}  77} &
  \cellcolor[HTML]{EFEFEF}{\color[HTML]{333333}  71} &
  \cellcolor[HTML]{EFEFEF}{\color[HTML]{333333}  73} &
  \cellcolor[HTML]{EFEFEF}{\color[HTML]{333333}   8} &
  \cellcolor[HTML]{9B9B9B}{\color[HTML]{333333} 0} &
  \cellcolor[HTML]{EFEFEF}{\color[HTML]{333333}  15} &
  \cellcolor[HTML]{FFFFFF}{\color[HTML]{333333}  26} &
  \cellcolor[HTML]{C0C0C0}{\color[HTML]{333333}   8} \\ \cline{2-2}
\cellcolor[HTML]{FFFFFF}{\color[HTML]{333333} } &
  {\color[HTML]{333333} RF+Fair-SMOTE} &
  \cellcolor[HTML]{FFFFFF}{\color[HTML]{333333}  58} &
  {\color[HTML]{333333}  79} &
  \cellcolor[HTML]{FFFFFF}{\color[HTML]{333333}  55} &
  \cellcolor[HTML]{FFFFFF}{\color[HTML]{333333}  65} &
  \cellcolor[HTML]{C0C0C0}{\color[HTML]{333333}   6} &
  \cellcolor[HTML]{C0C0C0}{\color[HTML]{333333}   6} &
  \cellcolor[HTML]{9B9B9B}{\color[HTML]{333333}   9} &
  \cellcolor[HTML]{EFEFEF}{\color[HTML]{333333}  18} &
  \cellcolor[HTML]{EFEFEF}{\color[HTML]{333333}  18} \\ \cline{2-2}
\cellcolor[HTML]{FFFFFF}{\color[HTML]{333333} } &
  {\color[HTML]{333333} RF+ROC} &
  \cellcolor[HTML]{FFFFFF}{\color[HTML]{333333}  60} &
  \cellcolor[HTML]{9B9B9B}{\color[HTML]{333333}  77} &
  \cellcolor[HTML]{FFFFFF}{\color[HTML]{333333}  56} &
  \cellcolor[HTML]{FFFFFF}{\color[HTML]{333333}  65} &
  \cellcolor[HTML]{C0C0C0}{\color[HTML]{333333}   5} &
  \cellcolor[HTML]{C0C0C0}{\color[HTML]{333333}   4} &
  \cellcolor[HTML]{9B9B9B}{\color[HTML]{333333}   8} &
  \cellcolor[HTML]{9B9B9B}{\color[HTML]{333333}   10} &
  \cellcolor[HTML]{FFFFFF}{\color[HTML]{333333}  44} \\ \cline{2-2}
\multirow{-5}{*}{\cellcolor[HTML]{FFFFFF}{\color[HTML]{333333} German: Sex}} &
  {\color[HTML]{333333} RF+FairMask} &
  {\color[HTML]{333333}  70} &
  \cellcolor[HTML]{C0C0C0}{\color[HTML]{333333}  73} &
  \cellcolor[HTML]{9B9B9B}{\color[HTML]{333333}  92} &
  {\color[HTML]{333333}  81} &
  \cellcolor[HTML]{C0C0C0}{\color[HTML]{333333}   5} &
  \cellcolor[HTML]{C0C0C0}{\color[HTML]{333333}   4} &
  \cellcolor[HTML]{9B9B9B}{\color[HTML]{333333}   8} &
  \cellcolor[HTML]{9B9B9B}{\color[HTML]{333333}   9} &
  \cellcolor[HTML]{9B9B9B}{\color[HTML]{333333} 0} \\ \hline
\cellcolor[HTML]{FFFFFF}{\color[HTML]{333333} } &
  {\color[HTML]{333333} RF} &
  {\color[HTML]{333333}  80} &
  {\color[HTML]{333333}  78} &
  \cellcolor[HTML]{9B9B9B}{\color[HTML]{333333}  82} &
  {\color[HTML]{333333}  80} &
  \cellcolor[HTML]{C0C0C0}{\color[HTML]{333333}   6} &
  \cellcolor[HTML]{EFEFEF}{\color[HTML]{333333}   9} &
  \cellcolor[HTML]{EFEFEF}{\color[HTML]{333333}  26} &
  \cellcolor[HTML]{EFEFEF}{\color[HTML]{333333}  55} &
  \cellcolor[HTML]{EFEFEF}{\color[HTML]{333333}  31} \\ \cline{2-2}
\cellcolor[HTML]{FFFFFF}{\color[HTML]{333333} } &
  {\color[HTML]{333333} RF+Random} &
  {\color[HTML]{333333}  80} &
  {\color[HTML]{333333}  77} &
  \cellcolor[HTML]{9B9B9B}{\color[HTML]{333333}  81} &
  {\color[HTML]{333333}  79} &
  \cellcolor[HTML]{EFEFEF}{\color[HTML]{333333}   7} &
  \cellcolor[HTML]{EFEFEF}{\color[HTML]{333333}  10} &
  \cellcolor[HTML]{9B9B9B}{\color[HTML]{333333}  10} &
  \cellcolor[HTML]{9B9B9B}{\color[HTML]{333333}  21} &
  \cellcolor[HTML]{9B9B9B}{\color[HTML]{333333} 0} \\ \cline{2-2}
\cellcolor[HTML]{FFFFFF}{\color[HTML]{333333} } &
  {\color[HTML]{333333} RF+Reweighing} &
  \cellcolor[HTML]{C0C0C0}{\color[HTML]{333333}  77} &
  \cellcolor[HTML]{C0C0C0}{\color[HTML]{333333}  74} &
  \cellcolor[HTML]{C0C0C0}{\color[HTML]{333333}  79} &
  \cellcolor[HTML]{C0C0C0}{\color[HTML]{333333}  76} &
  \cellcolor[HTML]{C0C0C0}{\color[HTML]{333333}   4} &
  \cellcolor[HTML]{9B9B9B}{\color[HTML]{333333}   2} &
  \cellcolor[HTML]{C0C0C0}{\color[HTML]{333333}  20} &
  \cellcolor[HTML]{C0C0C0}{\color[HTML]{333333}  40} &
  \cellcolor[HTML]{C0C0C0}{\color[HTML]{333333}   3} \\ \cline{2-2}
\cellcolor[HTML]{FFFFFF}{\color[HTML]{333333} } &
  {\color[HTML]{333333} RF+Fair-SMOTE} &
  {\color[HTML]{333333}  80} &
  {\color[HTML]{333333}  77} &
  \cellcolor[HTML]{9B9B9B}{\color[HTML]{333333}  82} &
  {\color[HTML]{333333}  80} &
  \cellcolor[HTML]{9B9B9B}{\color[HTML]{333333}   2} &
  \cellcolor[HTML]{C0C0C0}{\color[HTML]{333333}   6} &
  \cellcolor[HTML]{C0C0C0}{\color[HTML]{333333}  22} &
  \cellcolor[HTML]{C0C0C0}{\color[HTML]{333333}  44} &
  \cellcolor[HTML]{C0C0C0}{\color[HTML]{333333}  13} \\ \cline{2-2}
\cellcolor[HTML]{FFFFFF}{\color[HTML]{333333} } &
  {\color[HTML]{333333} RF+ROC} &
  \cellcolor[HTML]{C0C0C0}{\color[HTML]{333333}  78} &
  \cellcolor[HTML]{9B9B9B}{\color[HTML]{333333}  80} &
  \cellcolor[HTML]{EFEFEF}{\color[HTML]{333333}  73} &
  \cellcolor[HTML]{C0C0C0}{\color[HTML]{333333}  77} &
  \cellcolor[HTML]{C0C0C0}{\color[HTML]{333333}   4} &
  \cellcolor[HTML]{C0C0C0}{\color[HTML]{333333}   4} &
  \cellcolor[HTML]{EFEFEF}{\color[HTML]{333333}   25} &
  \cellcolor[HTML]{EFEFEF}{\color[HTML]{333333}   59} &
  \cellcolor[HTML]{FFFFFF}{\color[HTML]{333333}  69} \\ \cline{2-2}
\multirow{-5}{*}{\cellcolor[HTML]{FFFFFF}{\color[HTML]{333333} Bank: Age}} &
  {\color[HTML]{333333} RF+FairMask} &
  {\color[HTML]{333333}  80} &
  {\color[HTML]{333333}  78} &
  \cellcolor[HTML]{9B9B9B}{\color[HTML]{333333}  81} &
  {\color[HTML]{333333}  80} &
  \cellcolor[HTML]{C0C0C0}{\color[HTML]{333333}   5} &
  \cellcolor[HTML]{C0C0C0}{\color[HTML]{333333}   7} &
  \cellcolor[HTML]{9B9B9B}{\color[HTML]{333333}  11} &
  \cellcolor[HTML]{9B9B9B}{\color[HTML]{333333}  22} &
  \cellcolor[HTML]{9B9B9B}{\color[HTML]{333333} 0} \\ \hline
\cellcolor[HTML]{FFFFFF}{\color[HTML]{333333} } &
  {\color[HTML]{333333} RF} &
  {\color[HTML]{333333}  83} &
  {\color[HTML]{333333}  86} &
  \cellcolor[HTML]{C0C0C0}{\color[HTML]{333333}  78} &
  {\color[HTML]{333333}  81} &
  \cellcolor[HTML]{EFEFEF}{\color[HTML]{333333}  10} &
  \cellcolor[HTML]{C0C0C0}{\color[HTML]{333333}   6} &
  \cellcolor[HTML]{EFEFEF}{\color[HTML]{333333}  32} &
  \cellcolor[HTML]{EFEFEF}{\color[HTML]{333333}  55} &
  \cellcolor[HTML]{EFEFEF}{\color[HTML]{333333}   7} \\ \cline{2-2}
\cellcolor[HTML]{FFFFFF}{\color[HTML]{333333} } &
  {\color[HTML]{333333} RF+Random} &
  {\color[HTML]{333333}  83} &
  {\color[HTML]{333333}  85} &
  \cellcolor[HTML]{C0C0C0}{\color[HTML]{333333}  78} &
  {\color[HTML]{333333}  81} &
  \cellcolor[HTML]{C0C0C0}{\color[HTML]{333333}   8} &
  \cellcolor[HTML]{9B9B9B}{\color[HTML]{333333}   2} &
  \cellcolor[HTML]{9B9B9B}{\color[HTML]{333333}  24} &
  \cellcolor[HTML]{9B9B9B}{\color[HTML]{333333}  44} &
  \cellcolor[HTML]{C0C0C0}{\color[HTML]{333333}   3} \\ \cline{2-2}
\cellcolor[HTML]{FFFFFF}{\color[HTML]{333333} } &
  {\color[HTML]{333333} RF+Reweighing} &
  \cellcolor[HTML]{C0C0C0}{\color[HTML]{333333}  76} &
  \cellcolor[HTML]{C0C0C0}{\color[HTML]{333333}  71} &
  \cellcolor[HTML]{9B9B9B}{\color[HTML]{333333}  82} &
  \cellcolor[HTML]{C0C0C0}{\color[HTML]{333333}  75} &
  \cellcolor[HTML]{EFEFEF}{\color[HTML]{333333}  11} &
  \cellcolor[HTML]{EFEFEF}{\color[HTML]{333333}  11} &
  \cellcolor[HTML]{FFFFFF}{\color[HTML]{333333}  38} &
  \cellcolor[HTML]{EFEFEF}{\color[HTML]{333333}  54} &
  \cellcolor[HTML]{C0C0C0}{\color[HTML]{333333}   2} \\ \cline{2-2}
\cellcolor[HTML]{FFFFFF}{\color[HTML]{333333} } &
  {\color[HTML]{333333} RF+Fair-SMOTE} &
  {\color[HTML]{333333}  84} &
  {\color[HTML]{333333}  86} &
  \cellcolor[HTML]{C0C0C0}{\color[HTML]{333333}  79} &
  {\color[HTML]{333333}  82} &
  \cellcolor[HTML]{C0C0C0}{\color[HTML]{333333}   8} &
  \cellcolor[HTML]{9B9B9B}{\color[HTML]{333333}   4} &
  \cellcolor[HTML]{C0C0C0}{\color[HTML]{333333}  28} &
  \cellcolor[HTML]{C0C0C0}{\color[HTML]{333333}  48} &
  \cellcolor[HTML]{C0C0C0}{\color[HTML]{333333}   3} \\ \cline{2-2}
\cellcolor[HTML]{FFFFFF}{\color[HTML]{333333} } &
  {\color[HTML]{333333} RF+ROC} &
  \cellcolor[HTML]{C0C0C0}{\color[HTML]{333333}  75} &
  \cellcolor[HTML]{C0C0C0}{\color[HTML]{333333}  69} &
  \cellcolor[HTML]{C0C0C0}{\color[HTML]{333333}  77} &
  \cellcolor[HTML]{C0C0C0}{\color[HTML]{333333}  73} &
  \cellcolor[HTML]{EFEFEF}{\color[HTML]{333333}   10} &
  \cellcolor[HTML]{9B9B9B}{\color[HTML]{333333}   2} &
  \cellcolor[HTML]{9B9B9B}{\color[HTML]{333333}   23} &
  \cellcolor[HTML]{9B9B9B}{\color[HTML]{333333}   41} &
  \cellcolor[HTML]{FFFFFF}{\color[HTML]{333333}  97} \\ \cline{2-2}
\multirow{-5}{*}{\cellcolor[HTML]{FFFFFF}{\color[HTML]{333333} Health: Age}} &
  {\color[HTML]{333333} RF+FairMask} &
  {\color[HTML]{333333}  84} &
  {\color[HTML]{333333}  86} &
  \cellcolor[HTML]{C0C0C0}{\color[HTML]{333333}  79} &
  {\color[HTML]{333333}  82} &
  \cellcolor[HTML]{9B9B9B}{\color[HTML]{333333}   5} &
  \cellcolor[HTML]{9B9B9B}{\color[HTML]{333333}   4} &
  \cellcolor[HTML]{9B9B9B}{\color[HTML]{333333}  24} &
  \cellcolor[HTML]{9B9B9B}{\color[HTML]{333333}  43} &
  \cellcolor[HTML]{9B9B9B}{\color[HTML]{333333} 0} \\ \hline
\cellcolor[HTML]{FFFFFF}{\color[HTML]{333333} } &
  {\color[HTML]{333333} RF} &
  {\color[HTML]{333333}  82} &
  {\color[HTML]{333333}  66} &
  \cellcolor[HTML]{EFEFEF}{\color[HTML]{333333}  37} &
  \cellcolor[HTML]{C0C0C0}{\color[HTML]{333333}  47} &
  \cellcolor[HTML]{C0C0C0}{\color[HTML]{333333}   1} &
  \cellcolor[HTML]{C0C0C0}{\color[HTML]{333333}   2} &
  \cellcolor[HTML]{9B9B9B}{\color[HTML]{333333}   2} &
  \cellcolor[HTML]{EFEFEF}{\color[HTML]{333333}  20} &
  \cellcolor[HTML]{C0C0C0}{\color[HTML]{333333}   3} \\ \cline{2-2}
\cellcolor[HTML]{FFFFFF}{\color[HTML]{333333} } &
  {\color[HTML]{333333} RF+Random} &
  {\color[HTML]{333333}  81} &
  {\color[HTML]{333333}  65} &
  \cellcolor[HTML]{EFEFEF}{\color[HTML]{333333}  37} &
  \cellcolor[HTML]{C0C0C0}{\color[HTML]{333333}  47} &
  \cellcolor[HTML]{C0C0C0}{\color[HTML]{333333}   1} &
  \cellcolor[HTML]{C0C0C0}{\color[HTML]{333333}   2} &
  \cellcolor[HTML]{9B9B9B}{\color[HTML]{333333}   2} &
  \cellcolor[HTML]{EFEFEF}{\color[HTML]{333333}  18} &
  \cellcolor[HTML]{C0C0C0}{\color[HTML]{333333}   2} \\ \cline{2-2}
\cellcolor[HTML]{FFFFFF}{\color[HTML]{333333} } &
  {\color[HTML]{333333} RF+Reweighing} &
  \cellcolor[HTML]{EFEFEF}{\color[HTML]{333333}  42} &
  \cellcolor[HTML]{EFEFEF}{\color[HTML]{333333}  26} &
  \cellcolor[HTML]{9B9B9B}{\color[HTML]{333333}  88} &
  \cellcolor[HTML]{EFEFEF}{\color[HTML]{333333}  40} &
  \cellcolor[HTML]{9B9B9B}{\color[HTML]{333333} 0} &
  \cellcolor[HTML]{9B9B9B}{\color[HTML]{333333} 0} &
  \cellcolor[HTML]{9B9B9B}{\color[HTML]{333333}   2} &
  \cellcolor[HTML]{9B9B9B}{\color[HTML]{333333}   3} &
  \cellcolor[HTML]{9B9B9B}{\color[HTML]{333333} 0} \\ \cline{2-2}
\cellcolor[HTML]{FFFFFF}{\color[HTML]{333333} } &
  {\color[HTML]{333333} RF+Fair-SMOTE} &
  {\color[HTML]{333333}  82} &
  \cellcolor[HTML]{9B9B9B}{\color[HTML]{333333}  62} &
  \cellcolor[HTML]{C0C0C0}{\color[HTML]{333333}  42} &
  {\color[HTML]{333333}  50} &
  \cellcolor[HTML]{9B9B9B}{\color[HTML]{333333} 0} &
  \cellcolor[HTML]{C0C0C0}{\color[HTML]{333333}   2} &
  \cellcolor[HTML]{C0C0C0}{\color[HTML]{333333}   3} &
  \cellcolor[HTML]{C0C0C0}{\color[HTML]{333333}  15} &
  \cellcolor[HTML]{C0C0C0}{\color[HTML]{333333}   2} \\ \cline{2-2}
\cellcolor[HTML]{FFFFFF}{\color[HTML]{333333} } &
  {\color[HTML]{333333} RF+ROC} &
  \cellcolor[HTML]{C0C0C0}{\color[HTML]{333333}  77} &
  \cellcolor[HTML]{C0C0C0}{\color[HTML]{333333}  49} &
  \cellcolor[HTML]{C0C0C0}{\color[HTML]{333333}  57} &
  \cellcolor[HTML]{9B9B9B}{\color[HTML]{333333}  51} &
  \cellcolor[HTML]{C0C0C0}{\color[HTML]{333333}   0} &
  \cellcolor[HTML]{9B9B9B}{\color[HTML]{333333}   1} &
  \cellcolor[HTML]{C0C0C0}{\color[HTML]{333333}   3} &
  \cellcolor[HTML]{C0C0C0}{\color[HTML]{333333}   13} &
  \cellcolor[HTML]{FFFFFF}{\color[HTML]{333333}  23} \\ \cline{2-2}
\multirow{-5}{*}{\cellcolor[HTML]{FFFFFF}{\color[HTML]{333333} Default: Sex}} &
  {\color[HTML]{333333} RF+FairMask} &
  {\color[HTML]{333333}  82} &
  {\color[HTML]{333333}  65} &
  \cellcolor[HTML]{EFEFEF}{\color[HTML]{333333}  37} &
  \cellcolor[HTML]{C0C0C0}{\color[HTML]{333333}  47} &
  \cellcolor[HTML]{9B9B9B}{\color[HTML]{333333} 0} &
  \cellcolor[HTML]{9B9B9B}{\color[HTML]{333333} 0} &
  \cellcolor[HTML]{9B9B9B}{\color[HTML]{333333}   2} &
  \cellcolor[HTML]{C0C0C0}{\color[HTML]{333333}  16} &
  \cellcolor[HTML]{9B9B9B}{\color[HTML]{333333} 0} \\ \hline
\cellcolor[HTML]{FFFFFF}{\color[HTML]{333333} } &
  {\color[HTML]{333333} RF} &
  {\color[HTML]{333333}  87} &
  {\color[HTML]{333333}  66} &
  \cellcolor[HTML]{EFEFEF}{\color[HTML]{333333}  39} &
  \cellcolor[HTML]{C0C0C0}{\color[HTML]{333333}  49} &
  \cellcolor[HTML]{C0C0C0}{\color[HTML]{333333}   2} &
  \cellcolor[HTML]{EFEFEF}{\color[HTML]{333333}   7} &
  \cellcolor[HTML]{C0C0C0}{\color[HTML]{333333}   4} &
  \cellcolor[HTML]{EFEFEF}{\color[HTML]{333333}  33} &
  \cellcolor[HTML]{C0C0C0}{\color[HTML]{333333}   1} \\ \cline{2-2}
\cellcolor[HTML]{FFFFFF}{\color[HTML]{333333} } &
  {\color[HTML]{333333} RF+Random} &
  {\color[HTML]{333333}  86} &
  {\color[HTML]{333333}  65} &
  \cellcolor[HTML]{EFEFEF}{\color[HTML]{333333}  39} &
  \cellcolor[HTML]{C0C0C0}{\color[HTML]{333333}  49} &
  \cellcolor[HTML]{9B9B9B}{\color[HTML]{333333}   1} &
  \cellcolor[HTML]{9B9B9B}{\color[HTML]{333333}   3} &
  \cellcolor[HTML]{9B9B9B}{\color[HTML]{333333}   2} &
  \cellcolor[HTML]{C0C0C0}{\color[HTML]{333333}  23} &
  \cellcolor[HTML]{C0C0C0}{\color[HTML]{333333}   1} \\ \cline{2-2}
\cellcolor[HTML]{FFFFFF}{\color[HTML]{333333} } &
  {\color[HTML]{333333} RF+Reweighing} &
  \cellcolor[HTML]{C0C0C0}{\color[HTML]{333333}  76} &
  \cellcolor[HTML]{EFEFEF}{\color[HTML]{333333}  40} &
  \cellcolor[HTML]{9B9B9B}{\color[HTML]{333333}  76} &
  {\color[HTML]{333333}  52} &
  \cellcolor[HTML]{9B9B9B}{\color[HTML]{333333}   1} &
  \cellcolor[HTML]{C0C0C0}{\color[HTML]{333333}   5} &
  \cellcolor[HTML]{C0C0C0}{\color[HTML]{333333}   5} &
  \cellcolor[HTML]{9B9B9B}{\color[HTML]{333333}  14} &
  \cellcolor[HTML]{9B9B9B}{\color[HTML]{333333} 0} \\ \cline{2-2}
\cellcolor[HTML]{FFFFFF}{\color[HTML]{333333} } &
  {\color[HTML]{333333} RF+Fair-SMOTE} &
  {\color[HTML]{333333}  87} &
  \cellcolor[HTML]{C0C0C0}{\color[HTML]{333333}  64} &
  \cellcolor[HTML]{C0C0C0}{\color[HTML]{333333}  47} &
  {\color[HTML]{333333}  54} &
  \cellcolor[HTML]{C0C0C0}{\color[HTML]{333333}   2} &
  \cellcolor[HTML]{C0C0C0}{\color[HTML]{333333}   5} &
  \cellcolor[HTML]{C0C0C0}{\color[HTML]{333333}   4} &
  \cellcolor[HTML]{EFEFEF}{\color[HTML]{333333}  30} &
  \cellcolor[HTML]{C0C0C0}{\color[HTML]{333333}   2} \\ \cline{2-2}
\cellcolor[HTML]{FFFFFF}{\color[HTML]{333333} } &
  {\color[HTML]{333333} RF+ROC} &
  \cellcolor[HTML]{EFEFEF}{\color[HTML]{333333}  71} &
  \cellcolor[HTML]{EFEFEF}{\color[HTML]{333333}  35} &
  \cellcolor[HTML]{9B9B9B}{\color[HTML]{333333}  84} &
  \cellcolor[HTML]{C0C0C0}{\color[HTML]{333333}  49} &
  \cellcolor[HTML]{9B9B9B}{\color[HTML]{333333}   1} &
  \cellcolor[HTML]{C0C0C0}{\color[HTML]{333333}   5} &
  \cellcolor[HTML]{C0C0C0}{\color[HTML]{333333}   4} &
  \cellcolor[HTML]{9B9B9B}{\color[HTML]{333333}   13} &
  \cellcolor[HTML]{FFFFFF}{\color[HTML]{333333}  86} \\ \cline{2-2}
\multirow{-5}{*}{\cellcolor[HTML]{FFFFFF}{\color[HTML]{333333} MEPS: Race}} &
  {\color[HTML]{333333} RF+FairMask} &
  {\color[HTML]{333333}  87} &
  {\color[HTML]{333333}  67} &
  \cellcolor[HTML]{EFEFEF}{\color[HTML]{333333}  39} &
  \cellcolor[HTML]{C0C0C0}{\color[HTML]{333333}  49} &
  \cellcolor[HTML]{9B9B9B}{\color[HTML]{333333}   1} &
  \cellcolor[HTML]{9B9B9B}{\color[HTML]{333333}   3} &
  \cellcolor[HTML]{9B9B9B}{\color[HTML]{333333}   3} &
  \cellcolor[HTML]{C0C0C0}{\color[HTML]{333333}  25} &
  \cellcolor[HTML]{9B9B9B}{\color[HTML]{333333} 0} \\ \hline
\end{tabular}

\label{tab:result1}
\end{table*}

\begin{table*} 
\centering
\scriptsize
\caption{Summarized result of RQ2. Each cell is the mean rank across all datasets. Lower ranks are better and highlighted in darker colors. FairMask constantly obtains top ranks in all metrics.}
\begin{tabular}{|c|cccc|ccccc|}
\hline
 &
  Accuracy &
  Precision &
  Recall &
  F1 Score &
  AOD &
  EOD &
  SPD &
  DI &
  FR \\ \hline
RF &
  \cellcolor[HTML]{9B9B9B}1.1 &
  \cellcolor[HTML]{9B9B9B}1.2 &
  \cellcolor[HTML]{9B9B9B}1.9 &
  \cellcolor[HTML]{9B9B9B}1.2 &
  \cellcolor[HTML]{EFEFEF}2.3 &
  2.8 &
  \cellcolor[HTML]{EFEFEF}2.6 &
  3.0 &
  2.9 \\ \cline{1-1}
RF+Random &
  \cellcolor[HTML]{C0C0C0}1.2 &
  \cellcolor[HTML]{9B9B9B}1.2 &
  \cellcolor[HTML]{C0C0C0}2.2 &
  \cellcolor[HTML]{C0C0C0}1.6 &
  \cellcolor[HTML]{C0C0C0}1.6 &
  \cellcolor[HTML]{C0C0C0}1.7 &
  \cellcolor[HTML]{9B9B9B}1.6 &
  \cellcolor[HTML]{C0C0C0}2.3 &
  \cellcolor[HTML]{EFEFEF}2.4 \\ \cline{1-1}
RF+Reweighing &
  2.3 &
  \cellcolor[HTML]{EFEFEF}2.4 &
  \cellcolor[HTML]{9B9B9B}1.8 &
  \cellcolor[HTML]{EFEFEF}2.2 &
  \cellcolor[HTML]{C0C0C0}1.8 &
  \cellcolor[HTML]{C0C0C0}1.6 &
  \cellcolor[HTML]{C0C0C0}2.1 &
  \cellcolor[HTML]{EFEFEF}2.4 &
  \cellcolor[HTML]{C0C0C0}2.7 \\ \cline{1-1}
RF+Fair-SMOTE &
  \cellcolor[HTML]{EFEFEF}1.7 &
  \cellcolor[HTML]{C0C0C0}1.6 &
  \cellcolor[HTML]{9B9B9B}2.0 &
  \cellcolor[HTML]{9B9B9B}1.3 &
  \cellcolor[HTML]{C0C0C0}1.7 &
  \cellcolor[HTML]{EFEFEF}2.0 &
  \cellcolor[HTML]{C0C0C0}2.1 &
  \cellcolor[HTML]{C0C0C0}2.4 &
  \cellcolor[HTML]{EFEFEF}2.7 \\ \cline{1-1}
RF+ROC &
  \cellcolor[HTML]{FFFFFF}2.1 &
  \cellcolor[HTML]{EFEFEF}2.3 &
  \cellcolor[HTML]{9B9B9B}1.9 &
  \cellcolor[HTML]{EFEFEF}1.9 &
  \cellcolor[HTML]{C0C0C0}1.7 &
  \cellcolor[HTML]{EFEFEF}1.9 &
  \cellcolor[HTML]{C0C0C0}1.9 &
  \cellcolor[HTML]{9B9B9B}1.6 &
  \cellcolor[HTML]{FFFFFF}3.4 \\ \cline{1-1}
RF+FairMask &
  \cellcolor[HTML]{9B9B9B}1.1 &
  \cellcolor[HTML]{9B9B9B}1.3 &
  \cellcolor[HTML]{9B9B9B}1.9 &
  \cellcolor[HTML]{9B9B9B}1.2 &
  \cellcolor[HTML]{9B9B9B}1.3 &
  \cellcolor[HTML]{9B9B9B}1.4 &
  \cellcolor[HTML]{9B9B9B}1.3 &
  \cellcolor[HTML]{9B9B9B}1.8 &
  \cellcolor[HTML]{9B9B9B}1.0 \\ \hline
\end{tabular}
\label{tab:result3}
\end{table*}

\section{Results}\label{result} 

To assess the effectiveness of our proposed approach as compared to other benchmark methods, we design the experiment evaluation around 4 research questions (RQs).

\begin{figure}[!]
\centering
\includegraphics[width=.48\textwidth]{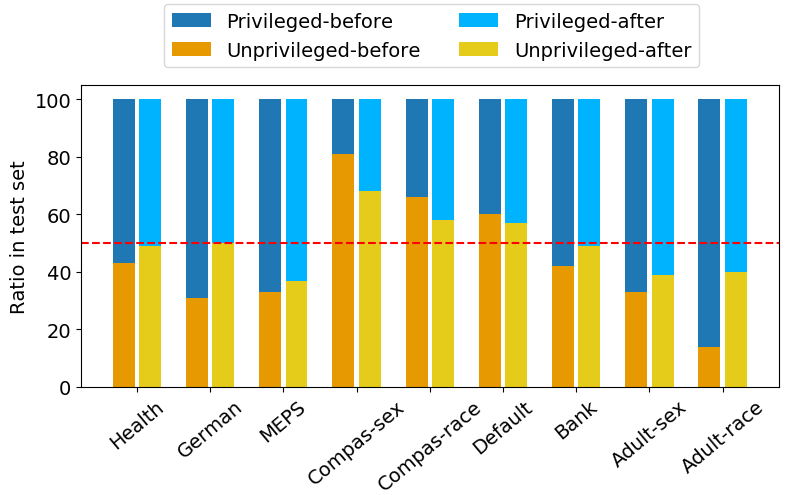}
\caption{  Result for RQ1. The ratio in percentage between the privileged and unprivileged groups before and after FairMask being applied.} 
\label{fig:result-rq1}
\end{figure}

\begin{blockquote}
\textbf{RQ1}: Does FairMask succeed in rebalancing the protected attributes?
\end{blockquote}
Before assessing the performance of FairMask, we would like to firstly check whether our method actually changes the distribution of protected attributes within the testing data.
Figure~\ref{fig:pos} and Figure~\ref{fig:neg} have shown the preliminary results, where the unprivileged group among the relabeled testing data has an increased possibility of receiving a favorable label, and the gap of the favorable label rates between the privileged and unprivileged group is reduced. 
Furthermore, in Figure~\ref{fig:result-rq1}, we compare the percentages of the privileged and unprivileged groups before/after FairMask being applied. 
The results are twofold: 
\begin{itemize}
    \item FairMask does change the distribution of protected attributes significantly\footnote{The percentages in the figure are mean values from 20 random repeats where the 2-tail paired t-test scores $<0.05$ in all cases.}.
    \item Even if the unprivileged group is not the (quantitatively) minority group, FairMask still achieves the rebalancing by shifting the percentage toward 50. In other words, FairMask is flexible and self-adaptive to various scenarios of unprivileged groups.
\end{itemize}
Our answer to RQ1 is: {\bf Yes, FairMask is not only effective but also adaptive in rebalancing the protected attributes.}

\begin{blockquote}
\textbf{RQ2}: How is the performance of our approach compared to other benchmarks, including state-of-the-art algorithms?
\end{blockquote}

Table~\ref{tab:result1} compares FairMask against other baselines.
Reweighing~\cite{kamiran2012data} is proposed by Kamiran et al. (introduced in \S\ref{bias-mitigation}) to mitigating bias via adjusting the instance weights for training samples in different groups (as determined by both their labels and protected attributes). Fair-SMOTE~\cite{Chakraborty2021BiasIM} is proposed by Chakraborty et al. to reduce bias by not only handling the data imbalance between target labels but also imbalanced distribution among different protected attributes. 
Reject Option based Classification (ROC) \cite{kamiran2012decision} is a post-processing method that optimizes for fairness by adjusting the classification threshold for the privileged and unprivileged group separately.
We chose them as our benchmark methods because (a) as a hybrid of pre-processing and post-processing, FairMask should be compared against other methods of the same category, and (b) all of them are supported in open-source packages and do not require any hyper-parameter tuning. \BLACK
In addition to these state-of-the-art methods, we also implement a "naive" baseline in our experiment, denoted as "Random", that randomly shuffles the protected attributes. 

Comparing FairMask against other baselines, we observe that the result is either (a)FairMask outperforms other methods in fairness metrics while maintaining the performance (in most cases, the default learner has top-ranked performance), or (b) FairMask reaches on-par fairness measures with some other baselines but obtains better performance at the same time. The summarized result is presented in Table~\ref{tab:result3}, where we can find that FairMask constantly obtains top ranks in both fairness and performance. It is also noteworthy that FairMask, by its design choices, can achieve perfect individual fairness while other baselines fail to improve it (even worsen in some cases).    

Thus, our answer to RQ2 is {\bf FairMask performs better or similar to the two state-of-the-art algorithms in terms of both fairness and performance.}

\begin{figure}[t!]
\centering
\includegraphics[width=.48\textwidth]{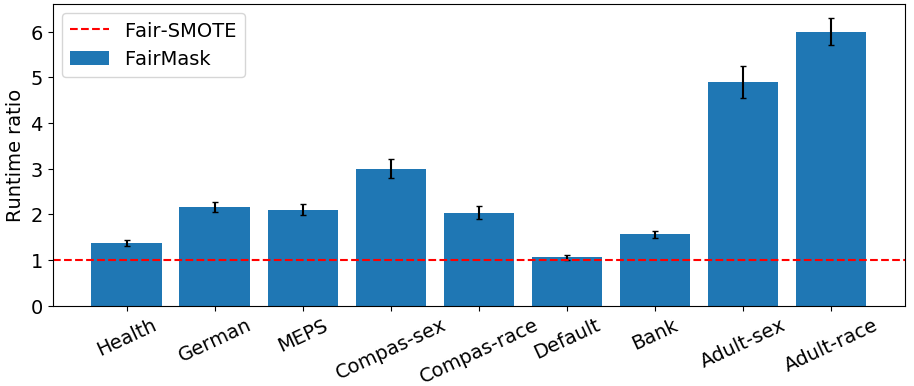}
\caption{Result for RQ3. The ratio is calculated by dividing the runtime of Fair-SMOTE over that of FairMask. A larger ratio means FairMask is faster. The datasets are sorted by the size in an ascending order.} 
\label{fig:result-rq4}
\end{figure}

\begin{blockquote}
\textbf{RQ3}: Is FairMask more scalable than Fair-SMOTE in terms of runtime complexity?
\end{blockquote}
 FairMask is built upon design choices that avoid synthetic data generation. This not only avoids the potential risk of introducing noise but also makes the whole framework more light-weighted. Figure~\ref{fig:result-rq4} presents the runtime of FairMask as compared to that of Fair-SMOTE on every dataset. While the datasets are sorted by their sizes, we do not see a proportional relationship between the size and runtime in either method. This could be because the runtime of the model is also influenced by the dimensionality of the training space. Moreover, since  Fair-SMOTE requires generating additional data, its runtime also depends on the extent of data imbalance: In cases where the data distribution is severely imbalanced, more synthetic data are required. 
Nevertheless, despite the variables described above, we can still observe apparent domination that FairMask runs constantly faster than Fair-SMOTE, which aligns with our design expectation. Thus, our answer to RQ3 is {\bf FairMask has a significantly shorter runtime, and therefore more scalable than Fair-SMOTE.}

\begin{table*}[t!]
\centering
\caption{Result for RQ4: FairMask is designed to be capable of mitigating bias on multiple protected attributes simultaneously. Similar to Table~\ref{tab:result1}, here cells with significantly better results are marked in a darker color.}
\begin{tabular}{|c|c|c|cccc|ccccc|}
\hline
Dataset &
  Method &
  \begin{tabular}[c]{@{}c@{}}Protected \\ Attribute\end{tabular} &
  Accuracy &
  Precision &
  Recall &
  F1 &
  AOD &
  EOD &
  SPD &
  DI &
  FR \\ \hline
 &
   &
  Sex &
  \cellcolor[HTML]{C0C0C0} &
  \cellcolor[HTML]{C0C0C0} &
   &
  \cellcolor[HTML]{C0C0C0} &
  8 &
  24 &
  18 &
  78 &
  20 \\ \cline{3-3}
 &
  \multirow{-2}{*}{RF} &
  Race &
  \multirow{-2}{*}{ \cellcolor[HTML]{C0C0C0}83} &
  \multirow{-2}{*}{\cellcolor[HTML]{C0C0C0}72} &
  \multirow{-2}{*}{53} &
  \multirow{-2}{*}{\cellcolor[HTML]{C0C0C0}61} &
  3 &
  10 &
  9 &
  49 &
  9 \\ \cline{2-12} 
  &
   &
  Sex &
   &
   & \cellcolor[HTML]{C0C0C0}
   &  
   &
  6 &
  21 &
  21 &
  54 &
  19 \\ \cline{3-3}
 &
  \multirow{-2}{*}{RF+Fair-SMOTE} &
  Race &
  \multirow{-2}{*}{75} &
  \multirow{-2}{*}{49} &
  \multirow{-2}{*}{ \cellcolor[HTML]{C0C0C0}71} &
  \multirow{-2}{*}{59} &
  3 &
  8 &
  10 &
  \cellcolor[HTML]{C0C0C0}32 &
  16 \\ \cline{2-12} 
 &
   &
  Sex &
   \cellcolor[HTML]{C0C0C0}&  
    \cellcolor[HTML]{C0C0C0}&
   &
   &
  \cellcolor[HTML]{C0C0C0}2 &
  \cellcolor[HTML]{C0C0C0}6 &
  \cellcolor[HTML]{C0C0C0}11 &
  \cellcolor[HTML]{C0C0C0}49 &
  \cellcolor[HTML]{C0C0C0}0 \\ \cline{3-3}
\multirow{-4}{*}{Adult} &
  \multirow{-2}{*}{RF+FairMask} &
  Race &
  \multirow{-2}{*}{ \cellcolor[HTML]{C0C0C0}83} &
  \multirow{-2}{*}{ \cellcolor[HTML]{C0C0C0}69} &
  \multirow{-2}{*}{52} &
  \multirow{-2}{*}{59} &
  \cellcolor[HTML]{C0C0C0}0 &
  \cellcolor[HTML]{C0C0C0}2 &
  \cellcolor[HTML]{C0C0C0}6 &
  \cellcolor[HTML]{C0C0C0}34 &
  \cellcolor[HTML]{C0C0C0}0 \\ \hline
 &
   &
  Sex &
   &
   &  \cellcolor[HTML]{C0C0C0}
   &
   &
  5 &
  10 &
  14 &
  19 &
  28 \\ \cline{3-3}
 &
  \multirow{-2}{*}{RF} &
  Race &
  \multirow{-2}{*}{65} &
  \multirow{-2}{*}{67} &
  \multirow{-2}{*}{ \cellcolor[HTML]{C0C0C0}73} &
  \multirow{-2}{*}{70} &
  2 &
  10 &
  14 &
  20 &
  24 \\ \cline{2-12} 
   &
   &
  Sex &
   &
   &
   &
   &
   \cellcolor[HTML]{C0C0C0}1 &
   \cellcolor[HTML]{C0C0C0}6 &
   \cellcolor[HTML]{C0C0C0}9 &
  18 &
  20 \\ \cline{3-3}
 &
  \multirow{-2}{*}{RF+Fair-SMOTE} &
  Race &
  \multirow{-2}{*}{65} &
  \multirow{-2}{*}{67} &
  \multirow{-2}{*}{70} &
  \multirow{-2}{*}{68} &
   3 &
   \cellcolor[HTML]{C0C0C0}7 &
   \cellcolor[HTML]{C0C0C0}13 &
  \cellcolor[HTML]{C0C0C0}17 &
  19 \\ \cline{2-12} 
 &
   &
  Sex &
   &
   &  \cellcolor[HTML]{C0C0C0}
   &
   &
  \cellcolor[HTML]{C0C0C0}1 &
  \cellcolor[HTML]{C0C0C0}5 &
  \cellcolor[HTML]{C0C0C0}9 &
  \cellcolor[HTML]{C0C0C0}14 &
  \cellcolor[HTML]{C0C0C0}0 \\ \cline{3-3}
\multirow{-4}{*}{Compas} &
  \multirow{-2}{*}{RF+FairMask} &
  Race &
  \multirow{-2}{*}{65} &
  \multirow{-2}{*}{66} &
  \multirow{-2}{*}{ \cellcolor[HTML]{C0C0C0}73} &
  \multirow{-2}{*}{69} &
  2 &
  \cellcolor[HTML]{C0C0C0}9 &
  \cellcolor[HTML]{C0C0C0}13 &
  \cellcolor[HTML]{C0C0C0}19 &
  \cellcolor[HTML]{C0C0C0}0 \\ \hline
\end{tabular}
\label{tab:result2}
\end{table*}

\begin{blockquote}
\textbf{RQ4}: Can FairMask handle multiple protected attributes?
\end{blockquote}
While under-represented in past research, it is a possible case scenario that a dataset contains more than one protected attributes in a dataset (just like Adult and Compas datasets). Fortunately, our design choices of FairMask makes itself extremely easy to be applied to cases with more than one protected attribute. 

To examine the effectiveness of FairMask, we conducted experiments on the Adult and Compas datasets, both of which contain two protected attributes: race and sex. Following our framework described in \ref{fig:frame}, we now need to build two extrapolation models for the two protected attributes respectively. After that, we will drop both protected attributes from the test data.
Since Fair-SMOTE is the only benchmark in this paper that is capable of handling multiple protected attributes, we only compare our approach against it.
As shown in Table~\ref{tab:result2}, compared to the default benchmark, FairMask can improve fairness in both protected attributes simultaneously while maintaining the predictive performance. When compared against Fair-SMOTE, FairMask still display better fairness while making less harm to performance.

It is also noteworthy that Fair-SMOTE uses the over-sampling tactic to reduce bias (in order to achieve balance among different combinations of protected and target attributes). Consequentially, the number of samples needed to over-sample explodes exponentially as the dimensionality of protected attributes increases, resulting in larger runtime complexity. FairMask, on the other hand, avoids this obstacle by introducing an extrapolation model rather than over-sampling the data. \BLACK 
In general, our answer to RQ4 is {\bf FairMask shows effectiveness in bias mitigation when handling multiple protected attributes simultaneously.}

\section{Threats to Validity}
\label{threat}

\noindent
\textbf{Sampling Bias:} While experimenting with other datasets may yield different results, we believe our extensive study here has shown the constant effectiveness of FairMask in various cases. Most of the prior works~\cite{NIPS2017_6988,chakraborty2019software,Kamiran:2018:ERO:3165328.3165686,Galhotra_2017,zhang2018mitigating} used one or two datasets where we used seven well-known datasets in our experiments. We have also observed other emerging datasets in the fairness fields, and we will try to extend our research scope once we verify the validity of the new datasets. In the future, we will explore more datasets and more learners. \\
\textbf{Evaluation Bias:} We used the five fairness metrics in this study, covering both definitions of group and individual fairness. Prior works \cite{Chakraborty_2020,10.1007/978-3-642-33486-3_3,hardt2016equality} used fewer metrics whereas IBM AIF360~\cite{bellamy2018ai} contains more than 50 metrics. More evaluation criteria will be examined in future work.\\
\textbf{Conclusion Validity:} Our experiments are based on the assumption that test data is unbiased and correctly labeled. Prior fairness studies also made the similar assumption~\cite{Chakraborty2021BiasIM,Biswas_2020,chakraborty2019software}.\\
\textbf{Internal Validity:} We used random forest model with mostly off-the-shelf parameters. However, hyper-parameters play a crucial role in the performance of ML models. Therefore, we cannot rule out the possibility that other ML models, after fine tuning, can achieve superior results. In the future, we will endeavor to address hyperparameter optimization for performance improvement. Moreover, our feature processing step during the experiment follows procedures found in prior works, especially those that are compared in this paper only in order to make a fair comparison. While other benchmark methods may have certain limitation in selecting features, our approach is actually applicable to all kinds of features.       \\
\textbf{External Validity:} Our work is limited to binary classification and tabular data which are very common in AI software. However, all the methods used in this paper can easily be extended in case of multi-class classification, and regression problems. In the future, we will try to extend our work to other domains of SE and ML.

\section{Discussion: Why FairMask?}\label{discussion}

In this section, we discuss what makes FairMask novel and distinguishable from prior works in this domain.

\subsection{Procedural Justice}\label{procedural}
In this paper, our experiment uses different measures to assess both group fairness (aod, eod, etc.) and individual fairness (flip rate). 
From the perspective of law practice, the notions of group fairness are more likely to reflect {\bf distributive justice}, which concerns fairness in terms of the distribution of rights~\cite{cook1983distributive} (in our case, the distribution of favorable labels among different social groups).
On the other hand, individual fairness tends to emphasize the importance of {\bf procedural justice}~\cite{tyler2002procedural,solum2004procedural}, which requires not only fair results but also transparency of the decision-making process such that ones can verify and monitor whether the procedure constantly guarantees fairness. The notions of procedural justice have recently gained more attention in the discussion of building fairer ML software~\cite{lee2019procedural,otting2018importance}.

In this paper, we claim that our design choice of improving individual fairness has made FairMask a suitable fit to satisfy procedural justice (as indicated by the flip rates). A model embedded with FairMask will not access the (actual) protected attributes during deployment time. However, we admit that FairMask is still not a perfect match according to the definition of procedural justice.
One of the significant reasons is model degradation. To prevent a model from compromised performance caused by distribution drift, ML models are recommended to be re-fitted periodically to adopt the updated data distribution. More importantly, in our case, we also need to update the extrapolation model within FairMask to better understand if the root cause of bias has also drifted (in different correlation with non-protected attributes).

Thus, while FairMask does not use the real protected attributes to make predictions on new data in deployment, those protected attributes are still collected in the background, such that we can still have access to the real protected attributes when re-training the prediction model. We notice that our design may not perfectly fulfill the requirement of procedural justice as we still collect protected information constantly. We believe FairMask has made a significant step toward procedural justice while there is still room for further improvement.
\BLACK

\subsection{Ethical Concerns}\label{ethical_concerns}
We can foresee the potential criticism that our approach might face: does FairMask mitigate bias or merely {\em hide} the bias since while the model no longer has access to protected attributes (e.g., gender, race), it still retains the influences of those attributes?
We say that this is demonstrably not true, and we would like to respond from two aspects. 

First of all, an important feature of the assessment methodology in this paper is that:
\begin{quote}
{\em When we assess the fairness extent of FairMask (in \S\ref{experiment}), that assessment uses all the protected attributes from the unaltered data.}
\end{quote}
Hence we can assert that our synthesis approach not only enables the procedural justice but it also reduces (but may not completely remove) other measurable effects of bias.

Moreover, there are many prior works in the literature that attempt to mitigate bias through modifying the protected attributes in their methods~\cite{kamiran2009classifying,feldman2015certifying,calmon2017optimized,calders2009building,kamiran2013techniques,zelaya2019parametrised}. For example, many pre-processing methods apply data "distortion/cleaning" on the training data, including the protected attributes, in order to fit a fairer prediction model on the modified. Other post-processing methods, which do not modify the protected attributes, directly change the prediction outcomes to mitigate bias. In short, the motivation and influence of FairMask and these prior works are essentially the same: Assuming or knowing a prediction model is unfair toward certain social groups, given an originally unfair outcome, we want to restore fairness by changing the undesired outcome. The only difference in FairMask is that we believe such undesired outcomes can be changed via properly modifying the protected attributes even after a biased model is trained.


\subsection{Why Does {\IT} Work?}\label{whywork} 
One frequently asked question is as follows. What is won by removing an attribute, then recreating its values via extrapolation from other attributes? Surely this extrapolation model just writes back the same values
that were removed?

In reply, we argue that the conclusions drawn
from the extrapolated data are actually different,
in certain small but crucial aspects, 
from the conclusions drawn from the raw data.
In FairMask, the relation between the protected and non-protected attributes is learned by an extrapolation model.
When new data instances arrive during the testing or deployment phase, FairMask generates synthetic values for protected attributes to mask actual values. In that approach, small variations in local data can be ``smoothed out'' by sampling across all the data through the extrapolation model. 
In this paper, Figure~\ref{fig:pos} and Figure~\ref{fig:neg} show that this kind of smoothing has a critical and significant effect on mitigating bias. Specifically, in those two figures, we look at the unfairness suffered by different social groupings:
\bi
\item In the test data, unprivileged groups have a much lower chance of receiving a favorable label while having a much higher chance of receiving an unfavorable label.
\item But when using our synthesized data generated from FairMask, that bias has been dramatically eliminated, leading the ratio toward an ideal equilibrium between the privileged and unprivileged groups.
\ei
We argue that biased decisions arise when a model
occasionally uses a protected attribute to make a decision while it has no need to. Our experience suggests that we can remove those ``occasional mistakes'', and thus mitigate bias.

\subsection{Does {\IT} Handle All Unfairness?}
When discussing this work with colleagues, we are often asked
if FairMask can mitigate against all the
potential injustices that might be created by AI.
In response, we say ``no''.
Mitigating the untoward effects of AI is a much broader problem than just exploring bias in algorithmic decision-making (as done in this paper).
The general problem of fairness is that influential groups in our society might mandate systems that (deliberately or unintentionally) disadvantage sub-groups within that society. A software system could satisfy all the metrics used to evaluate the extent of fairness
(as in Table~\ref{tab:metrics}) and still perpetuates social inequities. For example, (a)~software license fees might be so expensive that only a tiny monitory of organizations can boast they are ``fair''; or
(b)~the  skills required to use a model's API might be so elaborate that even if the model is fair, only an elite group of programmers can use it.
 
That said, as software developers, we cannot turn a blind eye to the detrimental social effects of our software.  While no single paper can hope to fix all social inequities, this paper shows how to improve the model involved in assessing one particular kind of unfairness (algorithmic decision-making bias).  
As to other kinds of fairness, they need to be explored and, hopefully,
research results like this one will motivate a larger community of researchers to take on the challenge of fairness.

\subsection{Explainable Extrapolation: Future Work}
Prior works either (a) do not offer interpretations on the cause of bias~\cite{creager2019flexibly,park2020readme,kamiran2012data}, or (b) offer instance-based summary~\cite{Chakraborty2021BiasIM} on the cause of bias. Although the latter one is human-comprehensible, we aim for generating more concise and structured interpretations via the extrapolation model.
As shown in \S\ref{explain}, we proposed an approach that explains the cause of bias in training data by extrapolating the correlation between the protected attributes and non-protected attributes. The experiment result in this paper has provided evidence that supports the presumption, which is that the privileged group may share a similar latent with the favorable-labeled group~\cite{creager2019flexibly,park2020readme}, as indicated by the explanations on both the target attribute (label) and the protected attribute. Such similarity within training data may mislead the classification model to wrongly emphasize the importance of the protected attribute, which is essentially a proxy of a combination of other informative attributes.

For future work, we look forward to formalizing the definition as well measurement for human comprehensibility level regarding group/individual fairness. For example, to verify a ML software to be fair in an "explainable" manner, one may need to provide user-centered experimental study with software stakeholders/users involved (e.g. Can a banker distinguish a de-biased mortgage application model from a biased one?).    While this paper lacks the human evaluation part to assert that our approach offers explainable bias mitigation, we believe our conjectures and concepts will inspire more promising work. 

\section{Conclusion} 

\label{conclusion}
Fairness in machine learning software has become a serious concern in the software engineering community.  
Many fairness methods synthesize  new samples~\cite{Chakraborty2021BiasIM,zelaya2019parametrised,kamiran2010classification} in order to better balance training data (expecting to remove certain biases). This paper tested an alternative approach, which assumes such synthesis might work better if it was 
{\em model-based} rather than {\em instance-based} 
(since the latter is more susceptible to minor variations in the data).

We found that we can endorse Chakraborty et al.~\cite{Chakraborty2021BiasIM} findings that bias might come from imbalanced data distributions. Moreover, instead of generating new data samples, we proposed FairMask, a better and faster approach that outperforms Fair-SMOTE. In addition, our approach also guarantees absolute procedural fairness. By avoiding using the actual protected attributes and synthesizing our own ones, our model can ensure that individuals that are only different in protected attributes will receive the same predictions. This is a significant improvement because, as revealed by our situational testing, sometimes such individual unfairness can exist among up to 20\% of the test data.
 
In experiments, we found that FairMask is performance-wise better (measured by fairness and performance metrics) than three state-of-the-art fairness algorithms. 
When looking at individual fairness (as indicated by the Flip Rates), FairMask can ensure perfect individual fairness while other benchmarks cannot.
Based on the above, we conclude that:
\begin{itemize}
    \item We can recommend FairMask for faster and more effective bias mitigation.
    \item FairMask greatly excludes the risk of individual unfairness: Two individuals who only differ in the protected attributes will always receive the same prediction outcomes. 
\end{itemize}

\section*{Acknowledgements}
This work was partially funded by 
a research grant from  Meta Inc and the Laboratory for Analytical Sciences, North Carolina State University,


\IEEEtriggeratref{56}
\bibliographystyle{IEEEtran}
\bibliography{main}

\end{document}

%% file: macros.tex
\usepackage{graphicx}
\usepackage{caption}
\usepackage{subcaption}
\usepackage{soul}
\usepackage{pgfplots}
\usepackage{epstopdf}
\usepackage{multirow}
\usepackage{listings}
\usepackage{fancybox}
\usepackage{color}
\usepackage{colortbl}
\usepackage{tcolorbox}
\makeatletter
\newcommand{\mybox}[1]{%
  \setbox0=\hbox{#1}%
  \setlength{\@tempdima}{\dimexpr\wd0+13pt}%
  \begin{tcolorbox}[boxrule=0.5pt, colback=white, arc=4pt,
	        left=6pt,right=6pt,top=6pt,bottom=6pt,boxsep=0pt]
			    #1
  \end{tcolorbox}
}

\definecolor{mygray}{gray}{.9}

\definecolor{mycolorhigh}{RGB}{168, 157, 89}
\definecolor{mycolormiddle}{RGB}{224, 224, 182}
\definecolor{mycolorlow}{RGB}{245, 245, 237}







%% file: main.bbl
\begin{thebibliography}{10}
\providecommand{\url}[1]{#1}
\csname url@samestyle\endcsname
\providecommand{\newblock}{\relax}
\providecommand{\bibinfo}[2]{#2}
\providecommand{\BIBentrySTDinterwordspacing}{\spaceskip=0pt\relax}
\providecommand{\BIBentryALTinterwordstretchfactor}{4}
\providecommand{\BIBentryALTinterwordspacing}{\spaceskip=\fontdimen2\font plus
\BIBentryALTinterwordstretchfactor\fontdimen3\font minus
  \fontdimen4\font\relax}
\providecommand{\BIBforeignlanguage}[2]{{%
\expandafter\ifx\csname l@#1\endcsname\relax
\typeout{** WARNING: IEEEtran.bst: No hyphenation pattern has been}%
\typeout{** loaded for the language `#1'. Using the pattern for}%
\typeout{** the default language instead.}%
\else
\language=\csname l@#1\endcsname
\fi
#2}}
\providecommand{\BIBdecl}{\relax}
\BIBdecl

\bibitem{DEFAULT}
\BIBentryALTinterwordspacing
``Uci:default of credit card clients data set,'' 2016. [Online]. Available:
  \url{https://archive.ics.uci.edu/ml/datasets/default+of+credit+card+clients}
\BIBentrySTDinterwordspacing

\bibitem{Chakraborty2021BiasIM}
\BIBentryALTinterwordspacing
J.~Chakraborty, S.~Majumder, and T.~Menzies, ``Bias in machine learning
  software: Why? how? what to do?'' in \emph{Proceedings of the 29th ACM Joint
  Meeting on European Software Engineering Conference and Symposium on the
  Foundations of Software Engineering}, ser. ESEC/FSE 2021.\hskip 1em plus
  0.5em minus 0.4em\relax New York, NY, USA: Association for Computing
  Machinery, 2021, p. 429–440. [Online]. Available:
  \url{https://doi.org/10.1145/3468264.3468537}
\BIBentrySTDinterwordspacing

\bibitem{tyler2002procedural}
T.~R. Tyler and E.~A. Lind, ``Procedural justice,'' in \emph{Handbook of
  justice research in law}.\hskip 1em plus 0.5em minus 0.4em\relax Springer,
  2002, pp. 65--92.

\bibitem{lee2019procedural}
M.~K. Lee, A.~Jain, H.~J. Cha, S.~Ojha, and D.~Kusbit, ``Procedural justice in
  algorithmic fairness: Leveraging transparency and outcome control for fair
  algorithmic mediation,'' \emph{Proceedings of the ACM on Human-Computer
  Interaction}, vol.~3, no. CSCW, pp. 1--26, 2019.

\bibitem{NIPS2017_6988}
\BIBentryALTinterwordspacing
F.~Calmon, D.~Wei, B.~Vinzamuri, K.~Natesan~Ramamurthy, and K.~R. Varshney,
  ``Optimized pre-processing for discrimination prevention,'' in \emph{Advances
  in Neural Information Processing Systems 30}, I.~Guyon, U.~V. Luxburg,
  S.~Bengio, H.~Wallach, R.~Fergus, S.~Vishwanathan, and R.~Garnett, Eds.\hskip
  1em plus 0.5em minus 0.4em\relax Curran Associates, Inc., 2017, pp.
  3992--4001. [Online]. Available:
  \url{http://papers.nips.cc/paper/6988-optimized-pre-processing-for-discrimination-prevention.pdf}
\BIBentrySTDinterwordspacing

\bibitem{mehrabi2021survey}
N.~Mehrabi, F.~Morstatter, N.~Saxena, K.~Lerman, and A.~Galstyan, ``A survey on
  bias and fairness in machine learning,'' \emph{ACM Computing Surveys (CSUR)},
  vol.~54, no.~6, pp. 1--35, 2021.

\bibitem{kamiran2012data}
F.~Kamiran and T.~Calders, ``Data preprocessing techniques for classification
  without discrimination,'' \emph{Knowledge and Information Systems}, vol.~33,
  no.~1, pp. 1--33, 2012.

\bibitem{kamiran2009classifying}
------, ``Classifying without discriminating,'' in \emph{2009 2nd international
  conference on computer, control and communication}.\hskip 1em plus 0.5em
  minus 0.4em\relax IEEE, 2009, pp. 1--6.

\bibitem{kilbertus2017avoiding}
N.~Kilbertus, M.~Rojas-Carulla, G.~Parascandolo, M.~Hardt, D.~Janzing, and
  B.~Sch{\"o}lkopf, ``Avoiding discrimination through causal reasoning,''
  \emph{arXiv preprint arXiv:1706.02744}, 2017.

\bibitem{zhang2016causal}
L.~Zhang, Y.~Wu, and X.~Wu, ``A causal framework for discovering and removing
  direct and indirect discrimination,'' \emph{arXiv preprint arXiv:1611.07509},
  2016.

\bibitem{ADULT}
\BIBentryALTinterwordspacing
``Uci:adult data set,'' 1994. [Online]. Available:
  \url{http://mlr.cs.umass.edu/ml/datasets/Adult}
\BIBentrySTDinterwordspacing

\bibitem{COMPAS}
\BIBentryALTinterwordspacing
``propublica/compas-analysis,'' 2015. [Online]. Available:
  \url{https://github.com/propublica/compas-analysis}
\BIBentrySTDinterwordspacing

\bibitem{GERMAN}
\BIBentryALTinterwordspacing
``Uci:statlog (german credit data) data set,'' 2000. [Online]. Available:
  \url{https://archive.ics.uci.edu/ml/datasets/Statlog+(German+Credit +Data)}
\BIBentrySTDinterwordspacing

\bibitem{BANK}
\BIBentryALTinterwordspacing
``Bank marketing uci,'' 2017. [Online]. Available:
  \url{https://www.kaggle.com/c/bank-marketing-uci}
\BIBentrySTDinterwordspacing

\bibitem{HEART}
\BIBentryALTinterwordspacing
``Uci:heart disease data set,'' 2001. [Online]. Available:
  \url{https://archive.ics.uci.edu/ml/datasets/Heart+Disease}
\BIBentrySTDinterwordspacing

\bibitem{MEPS}
\BIBentryALTinterwordspacing
``Medical expenditure panel survey,'' 2015. [Online]. Available:
  \url{https://meps.ahrq.gov/mepsweb/}
\BIBentrySTDinterwordspacing

\bibitem{feller2016computer}
A.~Feller, E.~Pierson, S.~Corbett-Davies, and S.~Goel, ``A computer program
  used for bail and sentencing decisions was labeled biased against blacks.
  it’s actually not that clear,'' \emph{The Washington Post}, vol.~17, 2016.

\bibitem{yeh2009comparisons}
I.-C. Yeh and C.-h. Lien, ``The comparisons of data mining techniques for the
  predictive accuracy of probability of default of credit card clients,''
  \emph{Expert Systems with Applications}, vol.~36, no.~2, pp. 2473--2480,
  2009.

\bibitem{shahriari2017ieee}
K.~Shahriari and M.~Shahriari, ``Ieee standard review—ethically aligned
  design: A vision for prioritizing human wellbeing with artificial
  intelligence and autonomous systems,'' in \emph{2017 IEEE Canada
  International Humanitarian Technology Conference (IHTC)}.\hskip 1em plus
  0.5em minus 0.4em\relax IEEE, 2017, pp. 197--201.

\bibitem{doi/10.2759/177365}
E.~Commission, C.~Directorate-General~for Communications~Networks, and
  Technology, \emph{Ethics guidelines for trustworthy AI}.\hskip 1em plus 0.5em
  minus 0.4em\relax Publications Office, 2019.

\bibitem{Fairness_Flow}
\BIBentryALTinterwordspacing
``Facebook says it has a tool to detect bias in its artificial intelligence,''
  2018. [Online]. Available:
  \url{https://qz.com/1268520/facebook-says-it-has-a-tool-to-detect-bias-in-its-artificial-intelligence/}
\BIBentrySTDinterwordspacing

\bibitem{FATE}
\BIBentryALTinterwordspacing
``Fate: Fairness, accountability, transparency, and ethics in ai,'' 2018.
  [Online]. Available:
  \url{https://www.microsoft.com/en-us/research/group/fate/}
\BIBentrySTDinterwordspacing

\bibitem{simonite2020google}
T.~Simonite, ``Google offers to help others with the tricky ethics of ai,''
  \emph{Ars Technica}, vol.~29, 2020.

\bibitem{FAT}
\BIBentryALTinterwordspacing
``Acm conference on fairness, accountability, and transparency (acm fat*).''
  [Online]. Available: \url{https://fatconference.org/}
\BIBentrySTDinterwordspacing

\bibitem{EXPLAIN}
\BIBentryALTinterwordspacing
``Explain 2019.'' [Online]. Available:
  \url{https://2019.ase-conferences.org/home/explain-2019}
\BIBentrySTDinterwordspacing

\bibitem{Chakraborty_2020}
\BIBentryALTinterwordspacing
J.~Chakraborty, S.~Majumder, Z.~Yu, and T.~Menzies, ``Fairway: A way to build
  fair ml software,'' in \emph{Proceedings of the 28th ACM Joint Meeting on
  European Software Engineering Conference and Symposium on the Foundations of
  Software Engineering}, ser. ESEC/FSE 2020.\hskip 1em plus 0.5em minus
  0.4em\relax New York, NY, USA: Association for Computing Machinery, 2020, p.
  654–665. [Online]. Available: \url{https://doi.org/10.1145/3368089.3409697}
\BIBentrySTDinterwordspacing

\bibitem{chakraborty2019software}
J.~Chakraborty, T.~Xia, F.~M. Fahid, and T.~Menzies, ``Software engineering for
  fairness: A case study with hyperparameter optimization,'' 2019.

\bibitem{9286091}
\BIBentryALTinterwordspacing
J.~Chakraborty, K.~Peng, and T.~Menzies, ``Making fair ml software using
  trustworthy explanation,'' in \emph{Proceedings of the 35th IEEE/ACM
  International Conference on Automated Software Engineering}, ser. ASE
  '20.\hskip 1em plus 0.5em minus 0.4em\relax New York, NY, USA: Association
  for Computing Machinery, 2020, p. 1229–1233. [Online]. Available:
  \url{https://doi.org/10.1145/3324884.3418932}
\BIBentrySTDinterwordspacing

\bibitem{Biswas_2020}
\BIBentryALTinterwordspacing
S.~Biswas and H.~Rajan, ``Do the machine learning models on a crowd sourced
  platform exhibit bias? an empirical study on model fairness,''
  \emph{Proceedings of the 28th ACM Joint Meeting on European Software
  Engineering Conference and Symposium on the Foundations of Software
  Engineering}, Nov 2020. [Online]. Available:
  \url{http://dx.doi.org/10.1145/3368089.3409704}
\BIBentrySTDinterwordspacing

\bibitem{Arvind}
A.~Narayanan, ``Tl;ds - 21 fairness definition and their politics by arvind
  narayanan,'' 2019.

\bibitem{feldman2015certifying}
M.~Feldman, S.~A. Friedler, J.~Moeller, C.~Scheidegger, and
  S.~Venkatasubramanian, ``Certifying and removing disparate impact,'' in
  \emph{proceedings of the 21th ACM SIGKDD international conference on
  knowledge discovery and data mining}, 2015, pp. 259--268.

\bibitem{grari2021fairness}
V.~Grari, S.~Lamprier, and M.~Detyniecki, ``Fairness without the sensitive
  attribute via causal variational autoencoder,'' \emph{arXiv preprint
  arXiv:2109.04999}, 2021.

\bibitem{10.1007/978-3-642-33486-3_3}
T.~Kamishima, S.~Akaho, H.~Asoh, and J.~Sakuma, ``Fairness-aware classifier
  with prejudice remover regularizer,'' in \emph{Machine Learning and Knowledge
  Discovery in Databases}, P.~A. Flach, T.~De~Bie, and N.~Cristianini,
  Eds.\hskip 1em plus 0.5em minus 0.4em\relax Berlin, Heidelberg: Springer
  Berlin Heidelberg, 2012, pp. 35--50.

\bibitem{oneto2019taking}
L.~Oneto, M.~Doninini, A.~Elders, and M.~Pontil, ``Taking advantage of
  multitask learning for fair classification,'' in \emph{Proceedings of the
  2019 AAAI/ACM Conference on AI, Ethics, and Society}, 2019, pp. 227--237.

\bibitem{chen2022maat}
Z.~Chen, J.~M. Zhang, F.~Sarro, and M.~Harman, ``Maat: A novel ensemble
  approach to addressing fairness and performance bugs for machine learning
  software,'' in \emph{Proceedings of the 30th ACM Joint European Software
  Engineering Conference and Symposium on the Foundations of Software
  Engineering (ESEC/FSE'22)}.\hskip 1em plus 0.5em minus 0.4em\relax ACM Press,
  2022.

\bibitem{kamiran2012decision}
F.~Kamiran, A.~Karim, and X.~Zhang, ``Decision theory for discrimination-aware
  classification,'' in \emph{2012 IEEE 12th International Conference on Data
  Mining}.\hskip 1em plus 0.5em minus 0.4em\relax IEEE, 2012, pp. 924--929.

\bibitem{Kamiran:2018:ERO:3165328.3165686}
F.~Kamiran, S.~Mansha, A.~Karim, and X.~Zhang, ``Exploiting reject option in
  classification for social discrimination control,'' \emph{Inf. Sci.}, 2018.

\bibitem{lundberg2017unified}
S.~M. Lundberg and S.-I. Lee, ``A unified approach to interpreting model
  predictions,'' in \emph{Advances in neural information processing systems},
  2017, pp. 4765--4774.

\bibitem{creager2019flexibly}
E.~Creager, D.~Madras, J.-H. Jacobsen, M.~Weis, K.~Swersky, T.~Pitassi, and
  R.~Zemel, ``Flexibly fair representation learning by disentanglement,'' in
  \emph{International conference on machine learning}.\hskip 1em plus 0.5em
  minus 0.4em\relax PMLR, 2019, pp. 1436--1445.

\bibitem{park2020readme}
S.~Park, D.~Kim, S.~Hwang, and H.~Byun, ``Readme: Representation learning by
  fairness-aware disentangling method,'' \emph{arXiv preprint
  arXiv:2007.03775}, 2020.

\bibitem{kamishima2011fairness}
T.~Kamishima, S.~Akaho, and J.~Sakuma, ``Fairness-aware learning through
  regularization approach,'' in \emph{2011 IEEE 11th International Conference
  on Data Mining Workshops}.\hskip 1em plus 0.5em minus 0.4em\relax IEEE, 2011,
  pp. 643--650.

\bibitem{vzliobaite2011handling}
I.~{\v{Z}}liobaite, F.~Kamiran, and T.~Calders, ``Handling conditional
  discrimination,'' in \emph{2011 IEEE 11th International Conference on Data
  Mining}.\hskip 1em plus 0.5em minus 0.4em\relax IEEE, 2011, pp. 992--1001.

\bibitem{Chawla:2002}
N.~V. Chawla, K.~W. Bowyer, L.~O. Hall, and W.~P. Kegelmeyer, ``Smote:
  Synthetic minority over-sampling technique,'' \emph{J. Artif. Int. Res.},
  2002.

\bibitem{bellamy2018ai}
R.~K. Bellamy, K.~Dey, M.~Hind, S.~C. Hoffman, S.~Houde, K.~Kannan, P.~Lohia,
  J.~Martino, S.~Mehta, A.~Mojsilovic \emph{et~al.}, ``Ai fairness 360: An
  extensible toolkit for detecting, understanding, and mitigating unwanted
  algorithmic bias,'' \emph{arXiv preprint arXiv:1810.01943}, 2018.

\bibitem{calmon2017optimized}
F.~P. Calmon, D.~Wei, B.~Vinzamuri, K.~N. Ramamurthy, and K.~R. Varshney,
  ``Optimized pre-processing for discrimination prevention,'' in
  \emph{Proceedings of the 31st International Conference on Neural Information
  Processing Systems}, 2017, pp. 3995--4004.

\bibitem{salazar2021automated}
R.~Salazar, F.~Neutatz, and Z.~Abedjan, ``Automated feature engineering for
  algorithmic fairness,'' \emph{Proceedings of the VLDB Endowment}, vol.~14,
  no.~9, pp. 1694--1702, 2021.

\bibitem{local_tse}
T.~Menzies, A.~Butcher, A.~Marcus, D.~Cok, F.~Shull, B.~Turhan, and
  T.~Zimmermann, ``Local versus global lessons for defect prediction and effort
  estimation,'' \emph{IEEE Transactions on software engineering}, vol.~39,
  no.~6, pp. 822--834, 2013.

\bibitem{menzies2014sharing}
T.~Menzies, E.~Kocaguneli, B.~Turhan, L.~Minku, and F.~Peters, \emph{Sharing
  data and models in software engineering}.\hskip 1em plus 0.5em minus
  0.4em\relax Morgan Kaufmann, 2014.

\bibitem{hardt2016equality}
M.~Hardt, E.~Price, and N.~Srebro, ``Equality of opportunity in supervised
  learning,'' 2016.

\bibitem{pleiss2017fairness}
G.~Pleiss, M.~Raghavan, F.~Wu, J.~Kleinberg, and K.~Q. Weinberger, ``On
  fairness and calibration,'' \emph{arXiv preprint arXiv:1709.02012}, 2017.

\bibitem{zhang2020white}
P.~Zhang, J.~Wang, J.~Sun, G.~Dong, X.~Wang, X.~Wang, J.~S. Dong, and T.~Dai,
  ``White-box fairness testing through adversarial sampling,'' in
  \emph{Proceedings of the ACM/IEEE 42nd International Conference on Software
  Engineering}, 2020, pp. 949--960.

\bibitem{mittas2012ranking}
N.~Mittas and L.~Angelis, ``Ranking and clustering software cost estimation
  models through a multiple comparisons algorithm,'' \emph{IEEE Transactions on
  software engineering}, vol.~39, no.~4, pp. 537--551, 2012.

\bibitem{xia2018hyperparameter}
T.~Xia, R.~Krishna, J.~Chen, G.~Mathew, X.~Shen, and T.~Menzies,
  ``Hyperparameter optimization for effort estimation,'' \emph{arXiv preprint
  arXiv:1805.00336}, 2018.

\bibitem{macbeth2011cliff}
G.~Macbeth, E.~Razumiejczyk, and R.~D. Ledesma, ``Cliff's delta calculator: A
  non-parametric effect size program for two groups of observations,''
  \emph{Universitas Psychologica}, vol.~10, no.~2, pp. 545--555, 2011.

\bibitem{hess2004robust}
M.~R. Hess and J.~D. Kromrey, ``Robust confidence intervals for effect sizes: A
  comparative study of cohen’sd and cliff’s delta under non-normality and
  heterogeneous variances,'' in \emph{annual meeting of the American
  Educational Research Association}, 2004, pp. 1--30.

\bibitem{Galhotra_2017}
\BIBentryALTinterwordspacing
S.~Galhotra, Y.~Brun, and A.~Meliou, ``Fairness testing: testing software for
  discrimination,'' \emph{Proceedings of the 2017 11th Joint Meeting on
  Foundations of Software Engineering - ESEC/FSE 2017}, 2017. [Online].
  Available: \url{http://dx.doi.org/10.1145/3106237.3106277}
\BIBentrySTDinterwordspacing

\bibitem{zhang2018mitigating}
B.~H. Zhang, B.~Lemoine, and M.~Mitchell, ``Mitigating unwanted biases with
  adversarial learning,'' 2018.

\bibitem{cook1983distributive}
K.~S. Cook and K.~A. Hegtvedt, ``Distributive justice, equity, and equality,''
  \emph{Annual review of sociology}, vol.~9, no.~1, pp. 217--241, 1983.

\bibitem{solum2004procedural}
L.~B. Solum, ``Procedural justice,'' \emph{S. CAl. l. reV.}, vol.~78, p. 181,
  2004.

\bibitem{otting2018importance}
S.~K. {\"O}tting and G.~W. Maier, ``The importance of procedural justice in
  human--machine interactions: Intelligent systems as new decision agents in
  organizations,'' \emph{Computers in Human Behavior}, vol.~89, pp. 27--39,
  2018.

\bibitem{calders2009building}
T.~Calders, F.~Kamiran, and M.~Pechenizkiy, ``Building classifiers with
  independency constraints,'' in \emph{2009 IEEE International Conference on
  Data Mining Workshops}.\hskip 1em plus 0.5em minus 0.4em\relax IEEE, 2009,
  pp. 13--18.

\bibitem{kamiran2013techniques}
F.~Kamiran, T.~Calders, and M.~Pechenizkiy, ``Techniques for
  discrimination-free predictive models,'' in \emph{Discrimination and privacy
  in the information society}.\hskip 1em plus 0.5em minus 0.4em\relax Springer,
  2013, pp. 223--239.

\bibitem{zelaya2019parametrised}
V.~Zelaya, P.~Missier, and D.~Prangle, ``Parametrised data sampling for
  fairness optimisation,'' \emph{KDD XAI}, 2019.

\bibitem{kamiran2010classification}
F.~Kamiran and T.~Calders, ``Classification with no discrimination by
  preferential sampling,'' in \emph{Proc. 19th Machine Learning Conf. Belgium
  and The Netherlands}.\hskip 1em plus 0.5em minus 0.4em\relax Citeseer, 2010,
  pp. 1--6.

\end{thebibliography}
